\documentclass[pdflatex,sn-mathphys-num]{sn-jnl}
\usepackage{graphicx}%
\usepackage{multirow}%
\usepackage{amsmath,amssymb,amsfonts}%
\usepackage{amsthm}%
\usepackage{mathrsfs}%
\usepackage[title]{appendix}%
\usepackage{colortbl}
\usepackage{xcolor}
\usepackage{textcomp}%
\usepackage{manyfoot}%
\usepackage{booktabs}%
\usepackage{algorithm}%
\usepackage{algorithmicx}%
\usepackage{algpseudocode}%
\usepackage{listings}%
\usepackage[table,xcdraw]{xcolor}
\usepackage{multirow}
\usepackage{multicol}
\usepackage{array}
\usepackage{makecell}

\theoremstyle{thmstyleone}%

\theoremstyle{thmstyletwo}%

\theoremstyle{thmstylethree}%

\begin{document}

\title[Better audio representations are more brain-like: linking model-brain alignment with performance in downstream auditory tasks]{Better audio representations are more brain-like: linking model-brain alignment with performance in downstream auditory tasks}

\author[1,2]{\fnm{Leonardo} \sur{Pepino}}\email{lpepino@dc.uba.ar}

\author[1,2]{\fnm{Pablo} \sur{Riera}}\email{priera@dc.uba.ar}

\author[1,2]{\fnm{Juan} \sur{Kamienkowski}}\email{juank@dc.uba.ar}

\author[1]{\fnm{Luciana} \sur{Ferrer}}\email{lferrer@dc.uba.ar}

\affil[1]{\orgdiv{Instituto de Investigación en Ciencias de la Computación (ICC)}, \orgname{CONICET-Universidad de Buenos Aires}, \orgaddress{\country{Argentina}}}

\affil[2]{\orgdiv{Departamento de Computación, FCEyN}, \orgname{Universidad de Buenos Aires (UBA)}, \orgaddress{\country{Argentina}}}

\abstract{Artificial neural networks are increasingly powerful models of brain computation, yet it remains unclear whether improving their performance in downstream tasks also makes their internal representations more similar to brain signals. To address this question in the auditory domain, we quantified the alignment between the internal representations of 36 different audio models and brain activity from two independent fMRI datasets. Using voxel-wise and component-wise regression, and representation similarity analysis, we found that recent self-supervised audio models with strong performance in diverse downstream tasks are better predictors of auditory cortex activity than previously studied models. 
To assess the quality of the audio representations, we evaluated these models in 6 auditory tasks from the HEAREval benchmark, spanning music, speech, and environmental sounds. This revealed strong positive Pearson correlations ($r>0.8$) between a model's overall task performance and its alignment with brain representations. Finally, we analyzed the evolution of the similarity between audio and brain representations during the pretraining of EnCodecMAE, a recent audio representation model. We discovered that brain similarity increases progressively and emerges early during pretraining, despite the model not being explicitly optimized for this objective. This suggests that brain-like representations can be an emergent byproduct of learning to reconstruct missing information from naturalistic audio data.}


\maketitle

\section{Introduction}\label{sec:intro}

Artificial Neural Networks (ANNs) are currently among the most promising models of brain computation, replicating several core properties of biological systems \cite{doerig2023neuroconnectionist}. As these models continue to excel at tasks traditionally associated with human intelligence, a central question arises: do their internal representations mirror those found in biological neural systems? Prior work has investigated this question across multiple domains, including vision \cite{gucclu2015deep, yamins2014performance, seeliger2018convolutional, eickenberg2017seeing, khaligh2014deep, cadena2019deep}, language \cite{awinstruction, goldstein2022correspondence, kumar2024shared, Schrimpf2020TheNAA}, and audition \cite{gucclu2016brains, khatami2020spiking, millet2021inductive, vaidya2022self, tuckute2023many}. These studies have shown that representations from deep neural networks (DNNs) can predict brain activity 
sometimes better than classical models designed to mimic human perception and cognition. A correspondence has also been observed between the hierarchical structure of DNNs and the organization of cortical processing stages. For example, Güçlü and van Gerven \cite{gucclu2015deep} found that early layers of a convolutional neural network (CNN) best predict activity in primary visual cortex (V1), while deeper layers more accurately predict responses in higher-level visual areas such as the lateral occipital complex, which is involved in object recognition. Similar correspondences have been reported for video \cite{eickenberg2017seeing, wen2018neural, sartzetaki2025one}, text \cite{mischler2024contextual, goldstein2025temporal, alkhamissi2025language} and auditory stimuli \cite{tuckute2023many, li2023dissecting, huang2018connecting}. For instance, \citet{tuckute2023many} examined the correspondence between deep audio models and fMRI responses in auditory cortex during naturalistic listening. They compared the activations of various pretrained audio models with brain responses through representational similarity analysis (RSA) and voxel activity regression, following the framework proposed by Doerig et al \cite{doerig2023neuroconnectionist}. Their findings demonstrated that many audio models, despite not being trained to approximate neural responses, exhibited strong alignment with brain activity.

Therefore, another question that arises is: as models get better at solving everyday tasks, do their representations also become more similar to our brain representations? Prior work  has shown that as language models are more capable of predicting the next word in a text, the alignment between their representations and those derived from brain activity increases \cite{Schrimpf2020TheNAA, Hong_2024, caucheteux2020language}. Further, aligning representations with brain data leads to better performance in downstream tasks \cite{Freteault2025AlignmentOA, li24l_interspeech}. These results support the Platonic Representation Hypothesis \cite{huh2024platonic}, which proposes that  models trained on different sensory modalities converge toward a shared, modality-agnostic “Platonic” representation of reality. \citet{huh2024platonic} showed empirical evidence for this hypothesis, by measuring the similarity between representations of models pretrained on text and images using different techniques like centered kernel alignment (CKA) and mutual \textit{k}-nearest neighbors (KNN), and observing that this similarity increases as models improve. A possible explanation suggested by the authors is that as models become more general and capable of solving diverse tasks, the space of representations that can simultaneously support these tasks becomes increasingly constrained\cite{huh2024platonic}. Consequently, since the tasks that the artificial systems are trained to optimize overlap with those that biological systems learn to solve, it is plausible that artificial and biological systems converge toward similar representations.

Here, we provide further evidence for this hypothesis by working in the auditory domain, focusing on deep neural networks trained for audio-related tasks, linking downstream performance with brain similarity for the first time in this domain. We aimed to answer the following research questions:

\vspace{1em}
\noindent\textbf{Are modern self-supervised audio models better aligned with brain than older models?} A few years ago, \citet{tuckute2023many}~studied the degree of similarity between audio model representations and brain representations \cite{tuckute2023many}. The models evaluated in that study included a diverse set of architectures, ranging from convolutional networks and recurrent models to transformers, and spanned tasks such as speech recognition, speaker separation, and audio captioning. However, those models were trained prior to 2022, with limited use of unsupervised objectives, and with data coming only from speech or environmental sounds. In this work, we update their analysis by incorporating recent state-of-the-art audio models: BEATs \cite{chen2023beats}, Dasheng \cite{dinkel2024dasheng}, and EnCodecMAE \cite{pepino25_interspeech}, which were trained using masked language modeling across diverse audio domains (speech, music and environmental sounds). Unlike most models studied in \cite{tuckute2023many}, these were trained in a fully self-supervised fashion, without fine-tuning on specific tasks. Additionally, we study models where a single hyperparameter or design choice was altered,  such as the training data, pretraining objectives and model size, measuring the impact of these factors on alignment.
\vspace{1em}

\noindent\textbf{How does the similarity with the brain evolve during pretraining?} We show that brain similarity increases progressively during the self-supervised pretraining of EnCodecMAE, despite not being explicitly involved in the optimization objective. This provides further evidence that alignment with the brain may emerge naturally when learning from naturalistic data and reconstructing missing information.
\vspace{1em}

\noindent\textbf{Do better audio models lead to more brain-like representations?} We measure the performance of each audio model in different audio downstream tasks, and compare the performance of each model with their similarity with the brain signals. Specifically, we show that models which perform better on a variety of downstream audio tasks like acoustic event detection and music genre classification, also exhibit stronger alignment with auditory cortical responses. This echoes similar findings in the language domain \cite{awinstruction} and suggests that optimizing for human-relevant tasks may promote brain-like representations. This finding provides evidence that the computational constraints of natural auditory processing may force different systems—whether biological or artificial—to converge upon a shared representation, as suggested by the Platonic Representation Hypothesis.
\vspace{1em}

\section{Results}\label{sec:results}

\begin{figure}
    \centering
    \includegraphics[width=\linewidth]{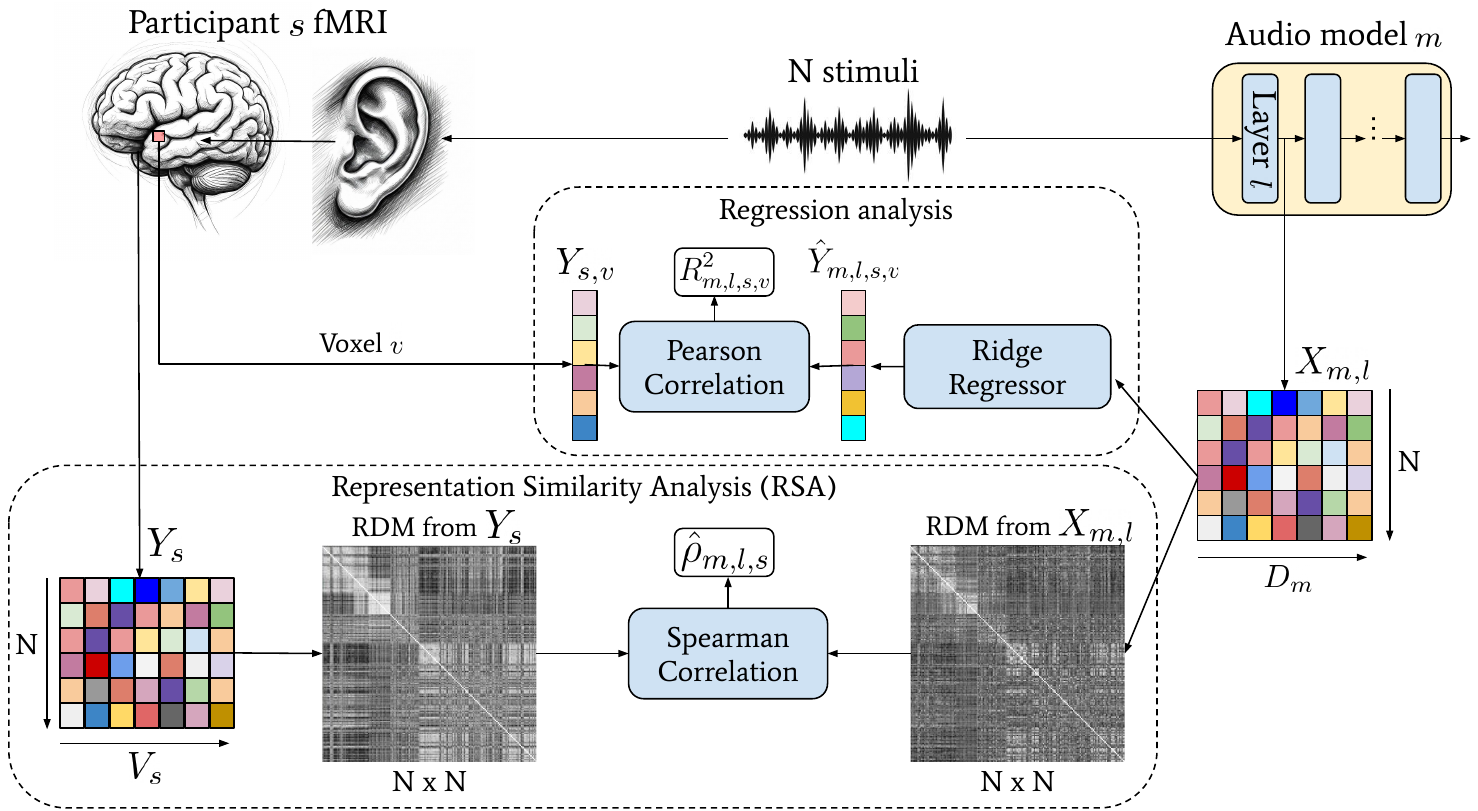}
    \caption{Schematic depicting the two main analysis we performed to compare audio and brain representations: regression analysis and representation similarity analysis (RSA). For regression, the target variable is the fMRI activity $Y_{s, v}$ of a voxel $v$ and subject $s$ and the predictor variable is the activation map from layer $l$ of an audio model $m$. For RSA, RDM matrices are calculated from $X_{m,l}$ and all the voxels from a subject $Y_{s}$ and compared using Spearman Correlation.}
    \label{fig:methods-summary}
\end{figure}

Figure \ref{fig:methods-summary} shows an overview of the analyses performed in this study. The main goal of these analyses is to get a measure of the similarity between audio and brain representations when presented with auditory stimuli. The two analysis techniques we used for measuring this similarity are regression and representation similarity. Briefly, in the regression analysis, for each voxel $v$ and subject $s$, an L2-regularized (ridge) linear regressor is trained to predict the summarized fMRI activity of $N=165$  auditory stimuli, collected into a vector \textbf{$Y_{s,v}$} of length $N$ corresponding to the activity in voxel $v$ of subject $s$. The inputs to the regressor consist of the audio representations of layer $l$ from model $m$, $X_{m,l}$ for the same $N$ stimuli. The output of the regressor is a vector of predictions $\hat{Y}_{m,l,s,v}$ which is compared with the target $Y_{s,v}$ through the Pearson determination coefficient, resulting in a matrix $R^2_{m,l,s,v}$. For each model $m$, voxel $v$ and subject $s$, the best layer $l$ and regularization weight $\alpha$ were searched using nested cross-validation on 5 equal-size splits balanced by stimuli type. This way, we obtain a matrix $R^2_{m,s,v}$ corresponding to the coefficient for the best layer for each model, voxel and subject. Then, we computed the median of $R^2_{m,s,v}$ across the voxels of each subject, obtaining $R^2_{m,s}$. Finally, we computed the mean across subjects, leading to a single summary metric, $R^2_m$, for how well model $m$ can predict brain activity. More details on the regression analysis can be found in Section \ref{sec:methods-regresion}.

In the Representation Similarity Analysis (RSA), for each subject $s$, the summarized fMRI activity for all stimuli and voxels is collected into a matrix $Y_s$ of size $N\times V_s$, where $V_s$ is the number of voxels for subject $s$. This representation is compared with the audio representation $X_{m,l}$ by computing representation dissimilarity matrices (RDMs) for both representations ($Y_s$ and $X_{m,l}$) and comparing them by calculating the Spearman correlation coefficient between their flattened lower triangular matrices to obtain a matrix $\rho_{m,l,s}$. The elements of the RDMs are calculated as one minus the Pearson correlation coefficient between every pair of stimuli. A single coefficient per model $m$ was finally obtained by averaging across subjects for each layer, and taking the maximum resulting value across layers. More details on the RSA analysis can be found in Section \ref{sec:methods-rsa}.

We evaluated the alignment between brain and model representations using two independent fMRI datasets: NH2015 \cite{norman2015distinct} and B2021 \cite{boebinger2021music}. These datasets capture BOLD responses in the auditory cortex of human participants as they listened to natural sounds. In each dataset, participants completed three scanning sessions where they heard the same two-second audio clips corresponding to everyday natural sounds ($N=165$). 
Details on fMRI acquisition and pre-processing are provided in Section \ref{sec:fmridata}.

Regarding the audio models, we analyzed those already studied in \citet{tuckute2023many} as well as more recent ones like BEATs \cite{chen2023beats}, Dasheng \cite{dinkel2024dasheng}, and EnCodecMAE \cite{pepino25_interspeech}. These newer models consist of a transformer encoder and were trained on large-scale unlabeled audio for masked language modeling (MLM), a task in which the model learns to reconstruct masked segments of the input from context. They have achieved strong performance across diverse tasks involving speech, music, and environmental sounds.

The main differences between the models are the targets they predict and datasets they use during pretraining.
BEATs (It 1) is pretrained on Audioset to predict quantized random projections of the input melspectrogram rectangular patches, while BEATs (It 2 and 3) replace these targets by discretized internal representations from the previous iteration (It 1 and 2 respectively). We also analyzed BEATs (FT), a version finetuned for acoustic event detection in Audioset.

EnCodecMAE (B) predicts discrete representations generated by EnCodec instead, a neural audio codec. As BEATs (It 2), EnCodecMAE (It 2) replaces this target by discretized internal representations. We also studied different EnCodecMAE sizes -- Large (L) and Small (S)--, different pretraining datasets-- Audioset (AS), LibriLight (LL), Free Music Archive (FMA), as well as a mixture of them which is used in the base model (B)--, and different input audio representations -- EnCodec features (EC), spectrograms (Spec), and melspectrograms which we use in the base model (B) --. We also analyzed a large version of EnCodecMAE pretrained using discrete representations from EnCodecMAE (It 2) as targets.

Finally, Dasheng is pretrained on a larger dataset (272K hours of diverse audio) compared to EnCodecMAE (B) which was trained with around 11K hours, and BEATs which was trained with around 5k hours. Differently from EnCodecMAE and BEATs, the targets are continuous and consist of the unmasked melspectrogram input representations. We analyzed 3 different sizes: B (86M parameters), XL (600M parameters) and XXL (1200M parameters), as well as a base version (B) finetuned for acoustic event detection in Audioset.
Full details on model configurations and feature extraction procedures are available in Section \ref{sec:models}.

\subsection{Are modern self-supervised audio models better aligned with brain signals than older models?} \label{sec:results-reg}

\begin{figure}
    \centering
    \includegraphics[width=\linewidth]{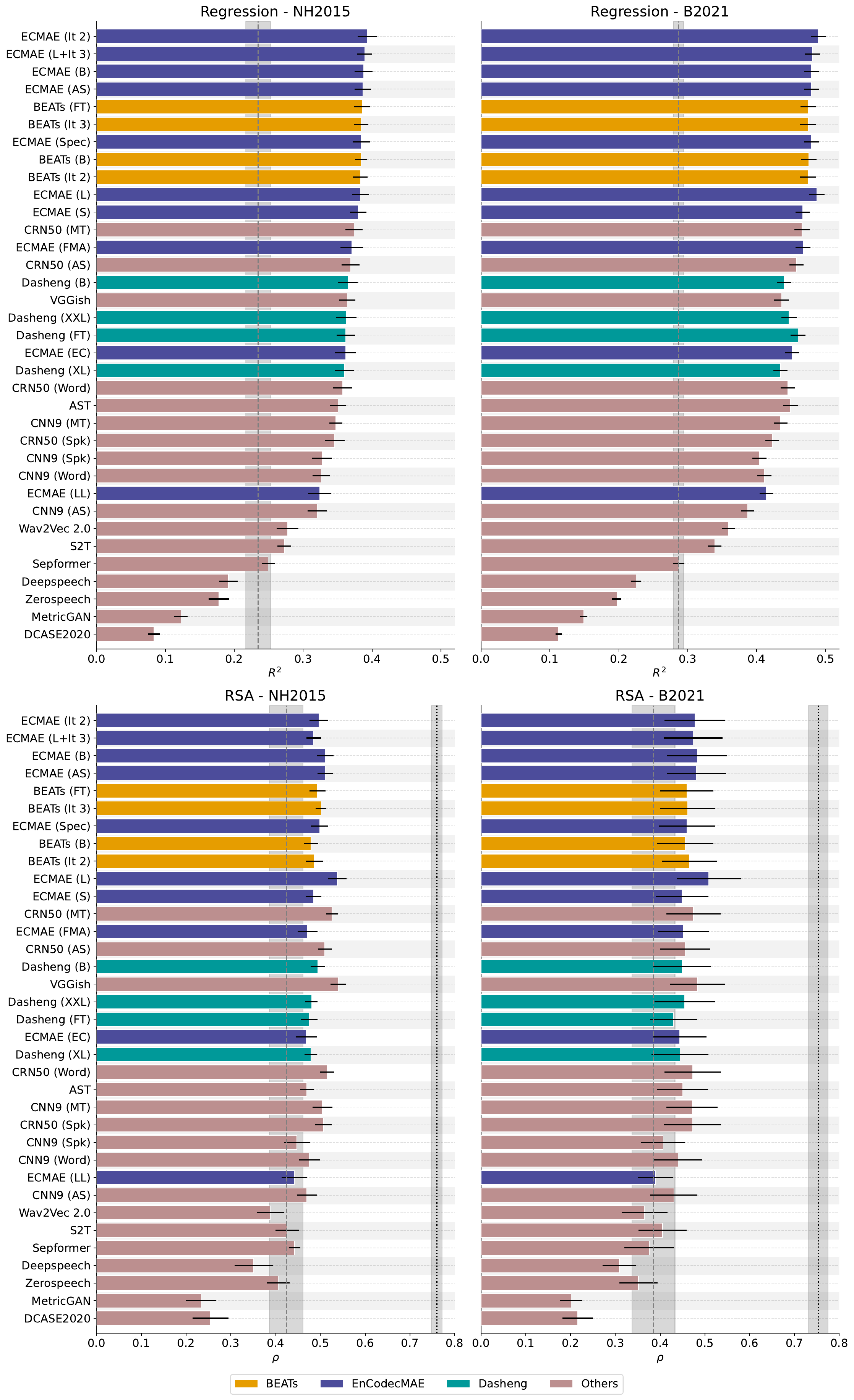}
    \caption{Top row: $R^2$ obtained for the analyzed audio models through the voxel-wise regression in the NH2015 (left column) and B2021 (right column). The gray line corresponds to the spectro-temporal baseline system. Bottom row: $\rho$ values obtained for the analyzed audio models through RSA. Gray lines correspond to the spectro-temporal baseline and the inter-subject RSA topline. Error bars reflect the standard error measured across subjects, and we ordered the model axis sorting by $R^2$ obtained in NH2015.}
    \label{fig:alignment}
\end{figure}

\subsubsection{Results from voxel regression}
In our first analysis, we compared the representations from 36 audio models in terms of their overall ability to predict fMRI representations using linear regression. 

Figure \ref{fig:alignment} (top row) shows the results obtained for the different models. The recent audio models pretrained with self-supervision in diverse audio (EnCodecMAE, BEATs and Dasheng), outperformed more specialized models studied in previous works (shown in pink bars).
This suggests that more recent models, which achieved stronger performance in general audio tasks, are also better predictors of the auditory cortex activity. Results show that ECMAE (LL), which is trained only on clean speech from LibriLight, exhibits a considerably lower alignment with brain representations than ECMAE (FMA), which is trained only on music from Free Music Archive, or ECMAE (AS) and ECMAE (B) which are pretrained on diverse audio spanning speech, music and environmental sounds. The results suggest that models trained on mixtures of diverse audio sources show the strongest brain activity prediction capabilities. In line with this observation, one possible explanation for Dasheng's relatively poorer alignment compared to the other recent models, despite being a high-performing model in downstream tasks, is that its pretraining dataset is composed mainly (96.6\%) of the ACAV100M dataset \cite{lee2021acav100m}. This dataset was constructed by selecting YouTube videos with high mutual information between audio and visual signals. As a result, events where sound and image are tightly coupled -- such as frontal speech or musical performances -- are overrepresented, while background sounds such as ambient noise, background speech, rain, or music may be underrepresented. This selection bias may result in a model that aligns less well to brain signals for being less representative of the data to which humans are commonly exposed. Altogether, these results highlight the key role of pretraining data on the alignment performance. 

Some recent works suggest that instruction finetuning \cite{aw2023instruction} or training models for specific tasks \cite{Aw2022TrainingLMA} lead to a better alignment between models and brain representations. In contrast with these prior works, we do not observe any significant difference between the checkpoints finetuned for acoustic event detection, BEATs (FT) and Dasheng (FT), and their corresponding base checkpoints that were not finetuned for a specific task, BEATs (It 3) and Dasheng (B). This suggests that the masked language modeling task already achieves representations that are aligned with the brain, without the need for training on a specific task with annotations. Finally, although it's been shown that the iterative refinement of targets in certain models like HuBERT plays an essential role in improving its representations and changes its organization across layers \cite{huo2025iterative}, we do not observe significant changes in the alignment to brain representations (ECMAE (L + It 3) vs ECMAE (L), ECMAE (It 2) vs ECMAE (B) and BEATs (It 2) and BEATs (It 3) vs BEATs (B).

\subsubsection{Similar results from RSA}
\begin{figure}[t]
    \centering
    \includegraphics[width=\linewidth]{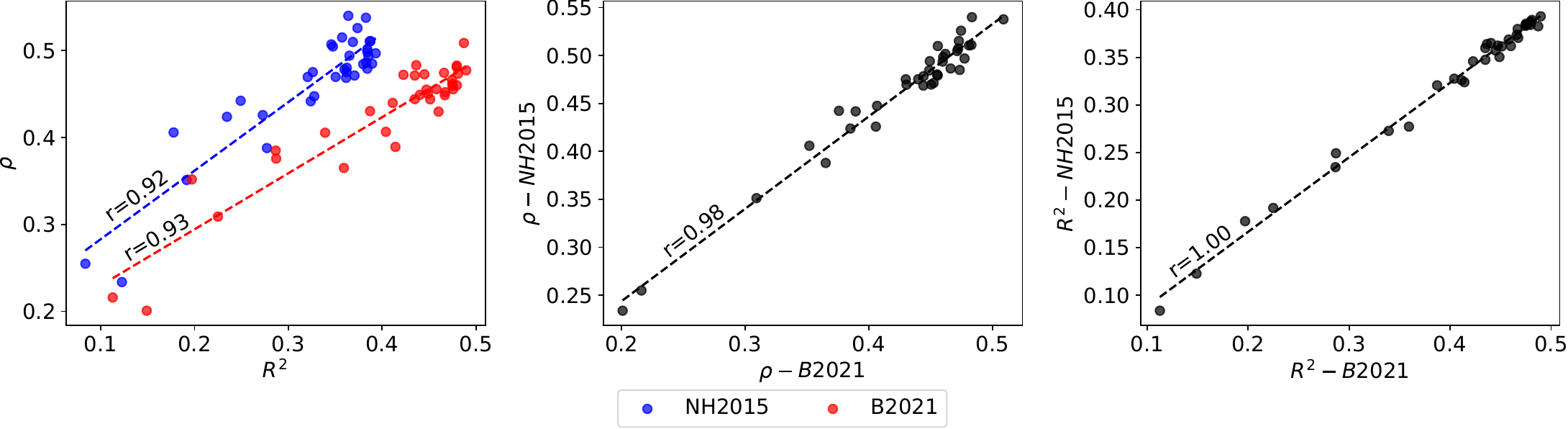}
    \caption{Left: Comparison of the scores obtained using voxel regression ($R^2$) and RSA ($\rho$) for both NH2015 and B2021 datasets. Center: Comparison of the $\rho$ scores obtained performing RSA on the B2021 and NH2015 datasets. Right: Comparison of the $R^2$ scores obtained performing voxel regression on the B2021 and NH2015 datasets.}
    \label{fig:compare-methods}
\end{figure}
Additionally, we performed a similarity analysis between representations from audio models and fMRI recordings using RSA (Fig. \ref{fig:methods-summary}), as explained in Section \ref{sec:results} and further detailed in Section \ref{sec:methods-rsa}. As seen in Figure \ref{fig:alignment} (bottom row), consistent with the regression analysis, pretraining with diverse data (ECMAE and ECMAE (AS)) leads to representations more similar to brain activity than models pretrained with domain-specific data like music (ECMAE (FMA)) and speech (ECMAE (LL)). Furthermore, we do not observe significant differences or clear trends between models that are finetuned or not, and between different iterations of target refinement, except for ECMAE (L) vs ECMAE (L + It. 3), where the refinement leads to a decrease in similarity. 

While, in line with the results from the regression analysis, representations from more recent models like EnCodecMAE and BEATs have the highest $\rho_m$ values, the difference with other models like VGGish or CochResNet50 is not as pronounced as in terms of $R^2$ values. This difference could be due to the fact that RSA cannot ignore dimensions in the model representations that are uncorrelated with brain activity. In contrast, regression analysis can find the subspace in the representation space which is relevant for brain activity prediction. In spite of these differences, both analyses yield similar conclusions and highly correlated scores (see Figure \ref{fig:compare-methods} - Left). Finally, Figure \ref{fig:compare-methods} (Center and Right) also shows that results are very similar for both datasets, despite involving different subjects.

\subsubsection{Recent models are better predictors of speech and music-related fMRI components}

In addition to the RSA and voxel-wise regression analysis, we performed a component-wise regression analysis, where the components are obtained as proposed in  \cite{norman2015distinct} by factorizing the average voxel activity matrix across subjects. Authors showed that six components accounted for 80\% of the variance. Moreover, these components were identified to be selective to low and high frequency tones (LF, HF), broadband spectra, tonal sounds (Pitch), speech, and music. 

We use these components as target vector instead of the individual voxel responses in the regression analysis. This approach has several advantages as it allows us to assess the degree to which models can predict brain activity, separately for each of the components, which are associated to different kinds of stimuli like speech, music or tonal sounds. It also makes the analysis more computationally efficient, as the number of target variables is reduced to six, instead of the number of voxels. A drawback of this analysis is that 20\% of the variance in the fMRI signal is not captured by the components, losing some potentially-valuable information. Another limitation is the loss of anatomical specificity compared to voxels, since components are not associated to a particular region in the brain. Further, components are obtained after pooling voxels across participants hence losing the ability to compute subject-specific metrics and variance across subjects.

\begin{figure}[H]
\centering
\includegraphics[width=\linewidth]{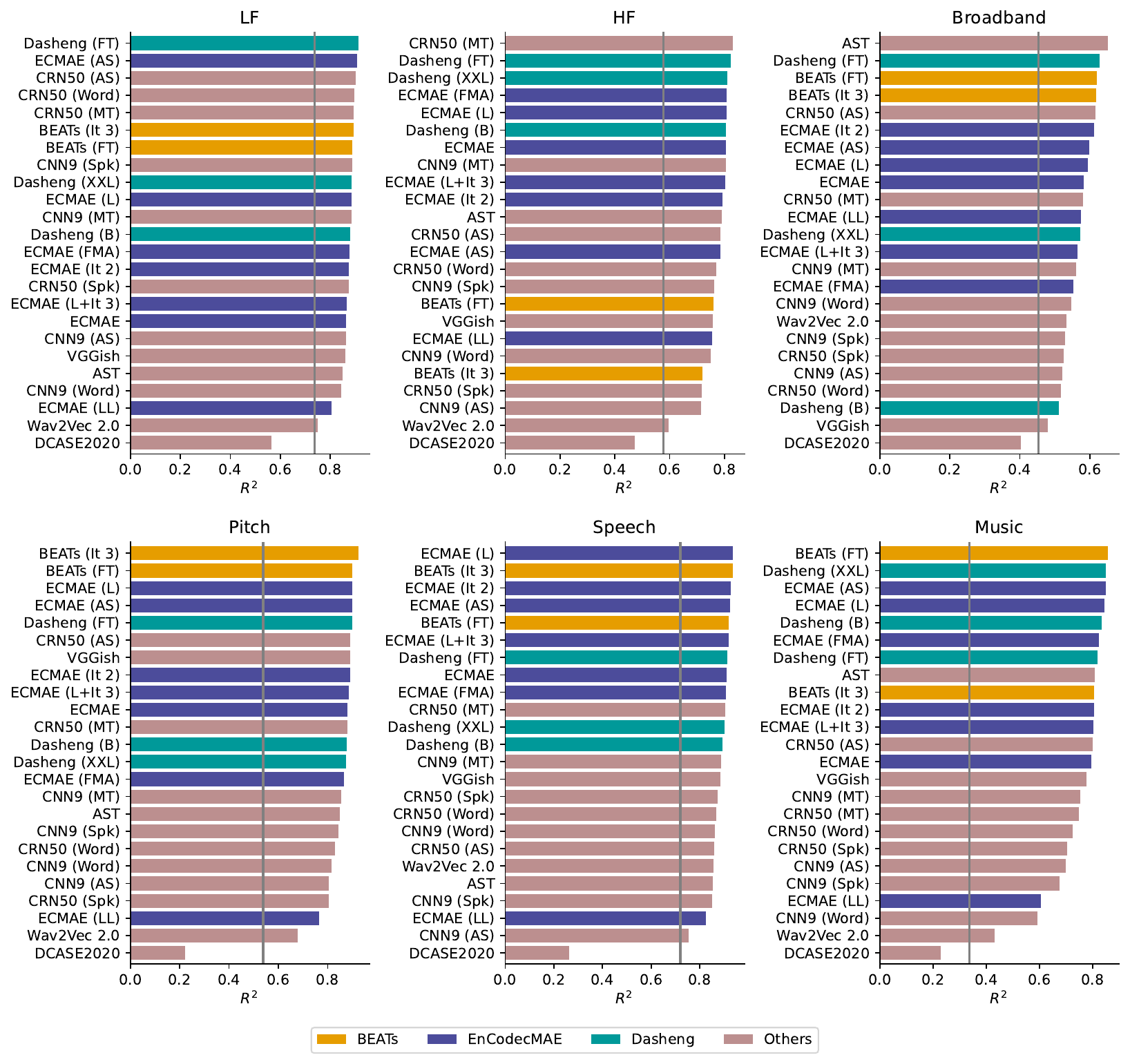}
\caption{$R^2$ values for each of the six components and a subset of the evaluated audio models (see Supplementary material for full plot). The gray lines correspond to the spectro-temporal baseline.}
\label{fig:pred-components}
\end{figure}

Figure~\ref{fig:pred-components} shows the performance of a subset of the audio models when predicting each of the six components. Interestingly, for the first components (related to spectral features such as low and high frequency energy), older models are good predictors, specially those that use cochleagram representations as inputs, like CRN50. 
In contrast, for components that are selective to speech and music stimuli, the most recent models introduced in this study, which are trained on larger and more diverse datasets using unsupervised objectives and transformer architectures, show superior performance.

Interestingly, contrary to what one might expect, models trained exclusively on speech or music are not the best predictors of the corresponding music and speech selective components. Instead, the best predictors are those trained on a combination of datasets.

\subsection{How does the similarity with the brain evolve during pretraining?}

\begin{figure}[h]
\centering
\includegraphics[width=\linewidth]{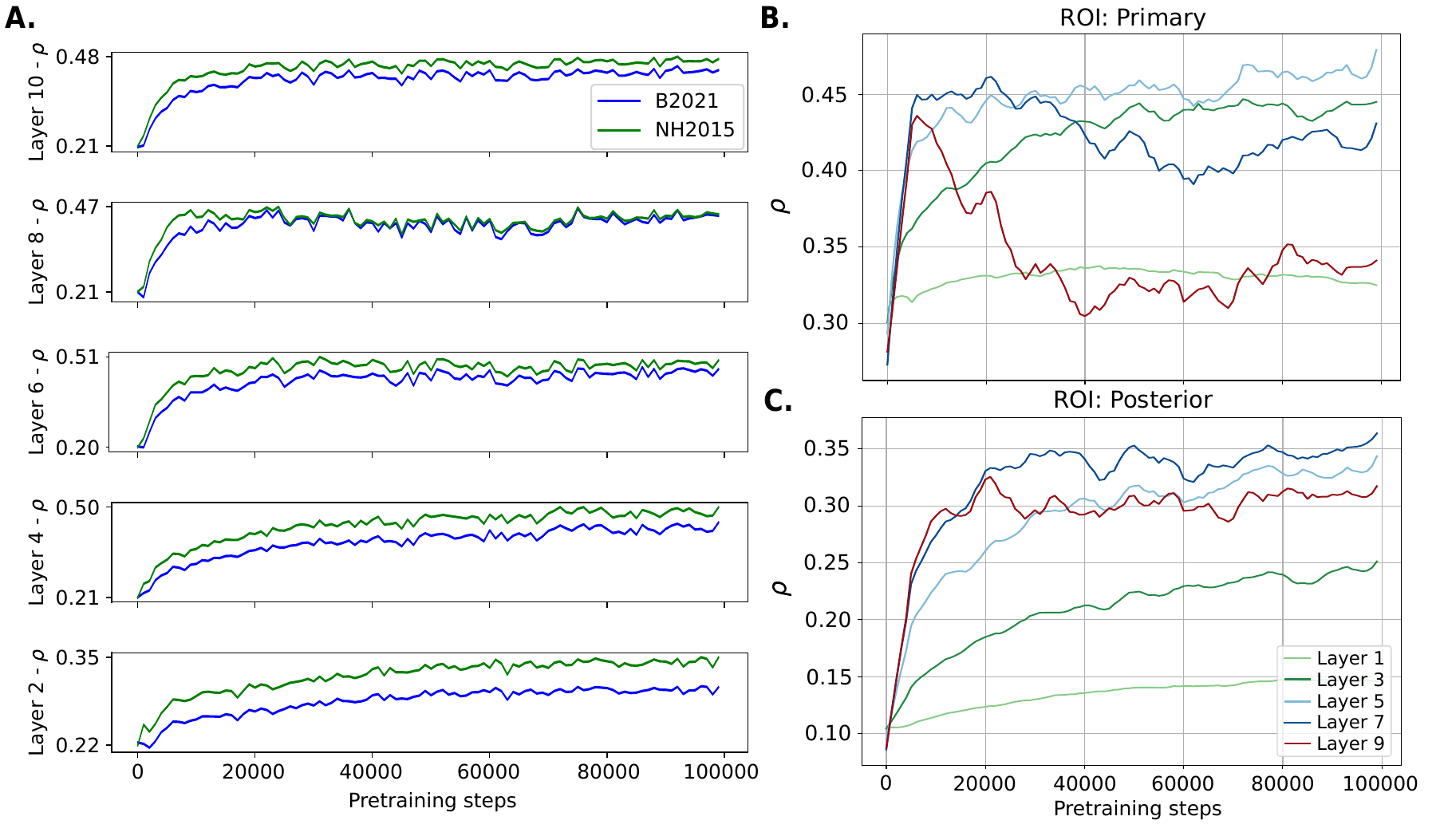}
\caption{A. Spearman's $\rho_{m,l}$ from the RSA analysis on the B2021 and NH2015 datasets for a subset of layers of EnCodecMAE throughout pretraining. B-C. Evolution in the first 100k pretraining steps of the similarity between audio and brain representations corresponding to the primary and posterior auditory regions (B and C respectively) in the NH2015 dataset, shown for each layer of the EnCodecMAE model. A Savitsky-Golay smoothing filter with window size=10 and order 3 was applied to the curves for visualization purposes.}
\label{fig:dynamics}
\end{figure}

An interesting question is whether optimizing the pretext task leads to the model’s representations becoming increasingly more aligned with those of the auditory cortex as the training process progresses. To investigate this, we analyze the Spearman correlation coefficient $\rho$ obtained with RSA for ECMAE (see Section \ref{sec:models}) as a function of the number of pretraining steps. If the pretext task naturally leads to greater alignment, we should observe an increase in similarity as pretraining progresses. Figure~\ref{fig:dynamics}A shows that the representations from different layers become increasingly similar to brain representations as pretraining progresses. It is important to note that during pretraining, there is no explicit optimization toward brain similarity, nor is any fMRI-based dataset used. The alignment, much like the strong downstream performance, is a byproduct of the model learning to reconstruct missing audio segments from context. 
Interestingly, alignment increases at a higher rate in the last layers compared with the first layers. Also, layers 4 and above achieve a higher similarity value than layer 2. 
A key finding is that alignment follows remarkably similar patterns across both datasets, despite involving different subjects, indicating the robustness of our methodology.

We also observed that structural differentiation emerges early on, mirroring patterns observed in the auditory cortex. This is illustrated in Figures~\ref{fig:dynamics}B and \ref{fig:dynamics}C, where layers 7 and 8 exhibit an early decrease in similarity when calculating RSA against only the voxels from the primary region, becoming less similar to primary auditory cortex representations than most other layers. In contrast, their similarity with the posterior region remains high and is among the highest across all layers.

Notably, the final layer does not follow the trend observed in earlier layers with respect to primary region similarity, maintaining a high similarity throughout training. We hypothesize that this is due to the pre-post normalization mechanism in ECMAE, which allows it to combine local information from earlier layers with global information from the later layers into the final layer. This incorporation of local information may in turn reduce its similarity with the posterior region compared to earlier layers (8 and 9). We do not show the evolution of the $\rho$ values during pretraining for the lateral and anterior regions since they exhibit patterns similar to that of the posterior region shown in Figure~\ref{fig:dynamics}C.

\subsection{Do better audio models lead to more brain-like representations?}
In addition to the brain alignment measurements, we measured the downstream performance of the different audio models, to determine whether there is a correlation between the representation quality and its alignment with auditory cortex. To measure downstream performance, we followed the HEAREval benchmark protocol \cite{turian2022hear}: Firstly, we evaluated in a subset of 6 HEAREval tasks encompassing music, speech and environmental sounds. The tasks are music note classification (NS), music genre classification (GC), speech commands recognition (SC), speech emotion recognition (ER), acoustic event detection (FSD) and acoustic event classification (ESC). These 6 tasks come from well-known datasets in the audio processing community and cover a diverse set of relevant auditory tasks. Secondly, we used the HEAREval downstream model, which consists of a multi-layer perceptron with limited hyperparameter exploration. Yet, unlike in the standard HEAREval approach where the last layer of the upstream model is used as input to the downstream model, here we combine the representations from all the layers, so that the procedure is better aligned with how the audio-brain alignment is measured. Finally, we obtain a summary metric by calculating z-scores from each task metric and then taking an average over the 6 tasks.

\begin{figure}
    \centering
    \includegraphics[width=\linewidth]{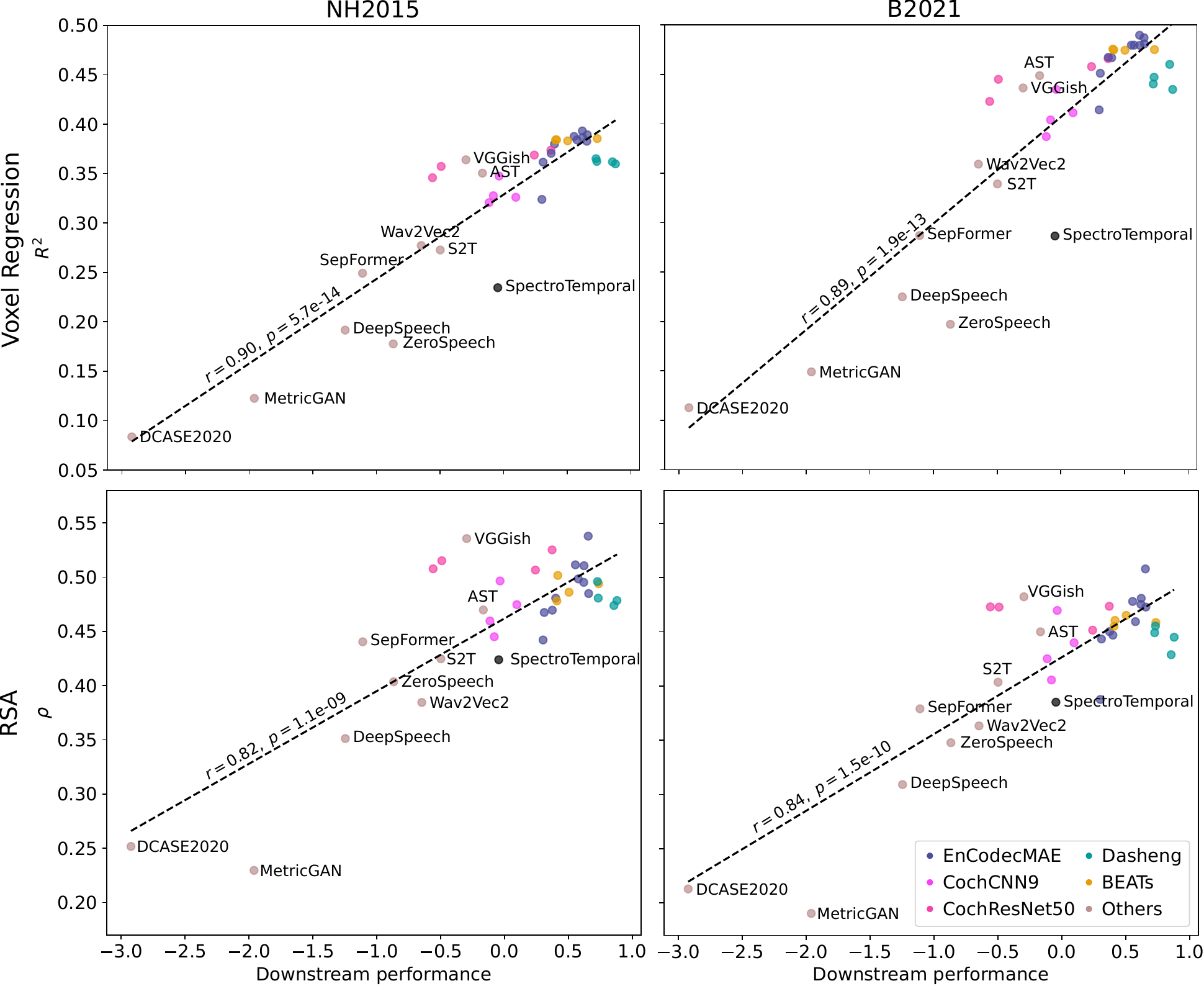}
    \caption{Overall downstream performance vs $R^2$ obtained from voxel regression and vs Spearman $\rho$ from RSA analysis in B2021 and NH2015 datasets for the 36 analysed models.}
    \label{fig:corr}
\end{figure}

The top row in Figure \ref{fig:corr} shows that the overall performance metric exhibits a strong positive Pearson correlation ($r=0.90$ and $r=0.89$ for NH2015 and B2021, respectively) with the $R^2$ from the regression analysis. Overall, there is a clear trend: models that perform better on downstream tasks also show stronger alignment with the auditory cortex. Even when excluding a few poorly performing models (DCASE2020, MetricGAN, DeepSpeech and SepFormer), the correlation remains high and significant ($r=0.71$ and $r=0.70$ for NH2015 and B2021, respectively).

The bottom row in Figure \ref{fig:corr} presents the same analysis using the $\rho$ value obtained with RSA. The conclusions remain consistent across both types of analysis, showing a clear positive correlation between downstream performance and alignment with the auditory cortex on both datasets.

We also analyzed the correlation between performance on each individual task and the $R^2$ coefficient from the regression analysis and the $\rho$ coefficient from RSA. Results are shown in Table \ref{tab:per-task-r2}, both including all models and filtering out those with an overall score below -1.0. When low-performing models are excluded, speech-related task performance (SC and ER) shows little to no correlation with model-brain alignment. The tasks most strongly correlated with alignment are those related to music genre classification (GC) and acoustic event classification (ESC) and detection (FSD). A possible explanation is that these tasks involve a broader and more diverse set of sounds, potentially engaging a wider range of brain activity patterns.

\begin{table}[]
\footnotesize
    \centering
\begin{tabular}{lll|c|c|c|c|c|c|c}
\toprule
 \multicolumn{3}{l|}{\multirow[c]{2}{*}{Method}} & \multicolumn{2}{c|}{Music} & \multicolumn{2}{c|}{Speech} & \multicolumn{2}{c|}{Env} & \\
& & & NS & GC & SC & ER & FSD & ESC & Overall \\
\midrule
\multirow[c]{2}{*}{Regression} & \multirow[c]{2}{*}{NH2015} & All & {\cellcolor{green!25}} .664 & {\cellcolor{green!25}} .819 & {\cellcolor{green!25}} .629 & {\cellcolor{green!25}} .631 & {\cellcolor{green!25}} .866 & {\cellcolor{green!25}} .866 & {\cellcolor{green!25}} \bfseries .902 \\
 &  & $> -1$ & {\cellcolor{yellow!25}} .392 & {\cellcolor{green!25}} .510 & {\cellcolor{red!25}} .217 & {\cellcolor{yellow!25}} .401 & {\cellcolor{green!25}} .653 & {\cellcolor{green!25}} .668 & {\cellcolor{green!25}} \bfseries .713 \\
 & \multirow[c]{2}{*}{B2021} & All & {\cellcolor{green!25}} .640 & {\cellcolor{green!25}} .800 & {\cellcolor{green!25}} .627 & {\cellcolor{green!25}} .633 & {\cellcolor{green!25}} .870 & {\cellcolor{green!25}} .868 & {\cellcolor{green!25}} \bfseries .895 \\
 & & $> -1$ & {\cellcolor{yellow!25}} .362 & {\cellcolor{green!25}} .468 & {\cellcolor{red!25}} .223 & {\cellcolor{yellow!25}} .405 & {\cellcolor{green!25}} .662 & {\cellcolor{green!25}} .676 & {\cellcolor{green!25}} \bfseries .703 \\
 \midrule
\multirow[c]{4}{*}{RSA} & \multirow[c]{2}{*}{B2021} & All & {\cellcolor{green!25}} .641 & {\cellcolor{green!25}} .823 & {\cellcolor{green!25}} .591 & {\cellcolor{green!25}} .526 & {\cellcolor{green!25}} .804 & {\cellcolor{green!25}} .795 & {\cellcolor{green!25}} \bfseries .842 \\
 &  & $> -1$ & {\cellcolor{yellow!25}} .370 & {\cellcolor{green!25}} .477 & {\cellcolor{red!25}} .088 & {\cellcolor{red!25}} .167 & {\cellcolor{green!25}} .459 & {\cellcolor{green!25}} .459 & {\cellcolor{green!25}} \bfseries .497 \\
 \cmidrule{2-10}
 & \multirow[c]{2}{*}{NH2015} & All & {\cellcolor{green!25}} .661 & {\cellcolor{green!25}} \bfseries .823 & {\cellcolor{green!25}} .537 & {\cellcolor{green!25}} .461 & {\cellcolor{green!25}} .781 & {\cellcolor{green!25}} .768 & {\cellcolor{green!25}} .813 \\
 &  & $> -1$ & {\cellcolor{green!25}} .455 & {\cellcolor{green!25}} \bfseries .503 & {\cellcolor{red!25}} -.016 & {\cellcolor{red!25}} .038 & {\cellcolor{yellow!25}} .427 & {\cellcolor{yellow!25}} .412 & {\cellcolor{yellow!25}} .437 \\
 \midrule
\multirow[c]{14}{*}{\thead{Regression \\ by \\Component}} & \multirow[c]{2}{*}{LF} & All & {\cellcolor{green!25}} \bfseries .807 & {\cellcolor{green!25}} .651 & {\cellcolor{yellow!25}} .424 & {\cellcolor{yellow!25}} .408 & {\cellcolor{green!25}} .746 & {\cellcolor{green!25}} .734 & {\cellcolor{green!25}} .760 \\
 &  & $> -1$ & {\cellcolor{green!25}} \bfseries .781 & {\cellcolor{yellow!25}} .368 & {\cellcolor{red!25}} .012 & {\cellcolor{red!25}} .119 & {\cellcolor{green!25}} .558 & {\cellcolor{green!25}} .528 & {\cellcolor{green!25}} .580 \\
 \cmidrule{2-10}
 & \multirow[c]{2}{*}{HF} & All & {\cellcolor{green!25}} .717 & {\cellcolor{green!25}} .788 & {\cellcolor{green!25}} .471 & {\cellcolor{green!25}} .512 & {\cellcolor{green!25}} .795 & {\cellcolor{green!25}} .791 & {\cellcolor{green!25}} \bfseries .821 \\
 &  & $> -1$ & {\cellcolor{green!25}} .587 & {\cellcolor{green!25}} .548 & {\cellcolor{red!25}} .020 & {\cellcolor{red!25}} .230 & {\cellcolor{green!25}} .569 & {\cellcolor{green!25}} .566 & {\cellcolor{green!25}} \bfseries .620 \\
 \cmidrule{2-10}
 & \multirow[c]{2}{*}{Broadband} & All & {\cellcolor{green!25}} .439 & {\cellcolor{green!25}} .670 & {\cellcolor{green!25}} .451 & {\cellcolor{green!25}} .584 & {\cellcolor{green!25}} \bfseries .851 & {\cellcolor{green!25}} .850 & {\cellcolor{green!25}} .775 \\
 &  & $> -1$ & {\cellcolor{red!25}} .149 & {\cellcolor{yellow!25}} .404 & {\cellcolor{red!25}} .073 & {\cellcolor{yellow!25}} .400 & {\cellcolor{green!25}} \bfseries .798 & {\cellcolor{green!25}} .770 & {\cellcolor{green!25}} .648 \\
 \cmidrule{2-10}
 & \multirow[c]{2}{*}{Pitch} & All & {\cellcolor{green!25}} .662 & {\cellcolor{green!25}} .840 & {\cellcolor{green!25}} .620 & {\cellcolor{green!25}} .597 & {\cellcolor{green!25}} .857 & {\cellcolor{green!25}} .862 & {\cellcolor{green!25}} \bfseries .894 \\
 &  & $> -1$ & {\cellcolor{red!25}} .266 & {\cellcolor{green!25}} .570 & {\cellcolor{red!25}} .112 & {\cellcolor{red!25}} .334 & {\cellcolor{green!25}} .657 & {\cellcolor{green!25}} \bfseries .665 & {\cellcolor{green!25}} .647 \\
 \cmidrule{2-10}
 & \multirow[c]{2}{*}{Speech} & All & {\cellcolor{green!25}} .575 & {\cellcolor{green!25}} .734 & {\cellcolor{green!25}} .670 & {\cellcolor{green!25}} .638 & {\cellcolor{green!25}} .807 & {\cellcolor{green!25}} .807 & {\cellcolor{green!25}} \bfseries .853 \\
 &  & $> -1$ & {\cellcolor{red!25}} .174 & {\cellcolor{red!25}} .295 & {\cellcolor{yellow!25}} .373 & {\cellcolor{yellow!25}} .447 & {\cellcolor{green!25}} .532 & {\cellcolor{green!25}} .560 & {\cellcolor{green!25}} \bfseries .611 \\
 \cmidrule{2-10}
 & \multirow[c]{2}{*}{Music} & All & {\cellcolor{green!25}} .571 & {\cellcolor{green!25}} .832 & {\cellcolor{green!25}} .527 & {\cellcolor{green!25}} .615 & {\cellcolor{green!25}} \bfseries .874 & {\cellcolor{green!25}} .860 & {\cellcolor{green!25}} .863 \\
 &  & $> -1$ & {\cellcolor{yellow!25}} .374 & {\cellcolor{green!25}} .609 & {\cellcolor{red!25}} .107 & {\cellcolor{yellow!25}} .396 & {\cellcolor{green!25}} .708 & {\cellcolor{green!25}} .709 & {\cellcolor{green!25}} \bfseries .724 \\

\bottomrule
\end{tabular}
    \caption{Pearson correlation between the downstream performance for each task and the alignment with the auditory cortex, as measured by voxel-wise and component-wise regression, and RSA. Results shown in green have a significance level $p<0.01$, in yellow $0.01 \leq p \leq 0.05$ and in red $p>0.05$. We also show the correlation values for a subset of models with an overall downstream performance higher than -1.}
    \label{tab:per-task-r2}
\end{table}

Table \ref{tab:per-task-r2} also shows the correlation between the downstream performance and the $R^2$ value for the different fMRI-derived components. 
In particular, we can see  that the alignment with the first two components (LF and HF), which reflect spectral attributes, correlates strongly with musical note classification performance, while others components show little or no significant correlation for the subset of better-performing systems. The alignment with the third component (Broadband), associated with impulsive (short and broadband) events, and the fourth component (Pitch), associated with temporally stable sounds, correlate best with the performance on environmental sound tasks. And the alignment of the last two components, associated with speech and music, correlate best with the environmental sound classification (ESC) and the overall performance metric.
On the other hand, the performance of speech-related tasks show the weakest correlation with alignment metrics on all components, which explains why models trained only on speech exhibit the worst alignment. Nevertheless, as expected, the highest correlation for these tasks corresponds to the speech-related component. Moreover, models that align best with the speech and music components are those that perform well across all tasks, as also shown in Figure \ref{fig:pred-components}, where the models introduced in this work are the ones that better aligned with these components.

Finally, the global metric and the FSD task are the only ones with highly significant positive correlations across all components (fully green columns). The FSD task requires detecting 200 heterogeneous sound classes, suggesting that brain alignment is strong for representations that can solve diverse tasks.

The results from these analyses support the Platonic Representation Hypothesis: as audio models improve, their alignment with other top-performing models (in our analysis, the brain) increases. 
Also, these findings suggest a practical implication for evaluating audio representations: given the low computational cost of performing representational similarity analysis with fMRI data, these methods could serve as alternatives or complements to benchmarks like HEAREval. For instance, during audio model pretraining, alignment with the auditory cortex could be monitored as a fast and efficient proxy for downstream performance.

\section{Discussion}\label{sec:discussion}
In this study, we analyzed the similarity between the representations from 36 different audio models and fMRI measurements by using both regularized linear regression and representational similarity analysis. Moreover, we measured the downstream performance of the audio models in a set of 6 auditory tasks spanning music, speech and environmental sound understanding. This set of measurements allowed us to establish a novel link between the quality of the audio representations, in terms of their usefulness to solve downstream auditory tasks, and their alignment with brain representations. For all the methods we used to measure brain alignment, and for both fMRI datasets, we observed strong positive Pearson correlations between the downstream performance and brain alignment.

We found that recent models such as ECMAE demonstrate a higher degree of alignment with brain representations compared to earlier models or traditional spectro-temporal baselines, indicating a trend toward biologically plausible representations. Models trained on more diverse datasets (e.g., AudioSet) exhibited greater alignment than those trained solely on speech, suggesting that data diversity during pretraining contributes positively to brain similarity. Our results regarding the correlation between downstream performance and brain alignment explain why these more recent audio models, that also have a good performance in downstream tasks, are better aligned with the brain.
Notably, performance on audio tasks such as acoustic event detection and genre music classification showed stronger correlations with brain similarity metrics, and task performance showed distinct correlation with the alignment for specific independent brain components. For instance, note classification performance correlated more strongly with the models alignment to frequency-selective components of primary auditory cortex, while acoustic event detection performance correlated with the alignment to components linked to broadband and tonal spectro-temporal features.

Finally, we showed that brain alignment is a characteristic that emerges early during the pretraining stage of ECMAE, in spite of not optimizing for this objective. These results support the Platonic Representation hypothesis, which suggests that as models get better they converge to a common representation. For example, vision and text representations, or in this case, audio and brain representations have higher similarity as the representations independently improve. One hypothesis for why this happens is that, as models are capable of solving more tasks, the space of possible solutions gets smaller, forcing different models to converge to similar representations, even if modalities are different, as they can be seen as different views of the same reality \cite{huh2024platonic}.

Our study is subject to some limitations, particularly regarding the fMRI data. Low temporal resolution of the fMRI measurements restricts our ability to assess fine-grained temporal encoding, which might be better captured using modalities such as EEG or MEG. 
The type of auditory stimuli that we used, which are daily natural sounds and span speech, music and environmental sounds, might favor models that were also pretrained with diverse sounds. 
Conversely, models trained solely on speech may appear less aligned because the fMRI stimuli included non-speech sounds that the model was not trained to represent. At the same time, 165 stimuli might not be enough to represent the wide range of sounds humans experience. For these reasons, the results and conclusions obtained in this work might depend on the used auditory stimuli set. Further studies should expand and diversify the stimuli set.

In addition, the correlation between downstream performance and brain similarity is limited by model coverage. Certain ranges of model performance are underrepresented (e.g., very low-performing models), which may affect correlation strength. Task selection and downstream model design also influence performance metrics and thus the interpretation of alignment. Particularly, using a set of 6 auditory tasks is not completely representative of the capacities of human audition. We expect correlation to become higher if we incorporate more auditory tasks for the overall downstream performance calculation. Also, it has been shown that the size and design of the downstream models can have a large impact on how different audio representations are ranked in terms of performance \cite{zaiem2025}. Despite these limitations, we have observed a clear trend in our analysis involving 36 diverse audio models.
Finally, linear regression models may filter out aspects of representations misaligned with brain data, while RSA, although holistic, is sensitive to transformations such as anisotropic scaling. Interestingly, despite these methodological differences, both analyses yielded consistent findings, lending robustness to our conclusions.

One interesting avenue for future work is to repeat our analyses using stimuli restricted to a specific domain, such as speech or music. Public datasets with fMRI recordings of naturalistic speech (e.g., storytelling) \cite{li2022petit, nastase2021narratives,lebel2021fmri} could facilitate a more focused evaluation of speech models and their neural alignment, which could be done with the source code released with this manuscript (see \url{https://github.com/mrpep/braindnn}). Moreover, the present study opens an opportunity to build stronger connections between neuroscience and machine learning, by using brain measurements to guide training of machine learning models. For example, one promising direction is to use fMRI-based RDMs to regularize model training or align audio representations with brain representations. Preliminary work in speech processing \cite{moussa2024improving, freteault2025alignment} suggests that aligning speech models with brain measurements can lead to performance improvements in different auditory tasks. Similar approaches were also explored in text processing \cite{toneva2019interpreting,negi2025braininformed}.

This research also invites exploration of the neuroconnectionist program beyond the human brain \cite{doerig2023neuroconnectionist}. Could models pretrained on animal vocalizations reveal insights about non-human auditory systems? Initial efforts in bird and animal sound modeling \cite{rauch2025can, moummad2024domain, hagiwara2023aves} hint at this possibility, but a more explicit comparison between animal-trained models and animal brain activity remains an exciting direction for future research.

Finally, one important question that emerges from this work is whether alignment with human brain representations is necessary for good downstream performance. Could there exist high-performing models that do not resemble brain activity? Identifying such counterexamples—models with low neural similarity but good task performance, or vice versa—could provide insight into the uniqueness of human-like representations. Also, finding such counter-examples would bring evidence that the platonic representation hypothesis \cite{huh2024platonic} might be incorrect.

\section{Methods}\label{sec:methods}

\subsection{fMRI datasets and preprocessing}\label{sec:fmridata}

We used the following two fMRI datasets \cite{norman2015distinct,boebinger2021music,tuckute2023many}:

\begin{itemize}
    \item NH2015 \cite{norman2015distinct}: This dataset contains fMRI recordings from 8 participants without musical training, all native English speakers aged 19–25 years. During scanning, participants listened to $N=165$ auditory stimuli, each 2 seconds in duration. The stimuli consisted of everyday sounds, presented in blocks in which each stimulus was repeated 5 times. Each participant completed 3 sessions of approximately 90 minutes, with each session divided into 11 runs comprising 15 stimulus blocks and 4 silent blocks. To monitor attention, one repetition of each stimulus was presented at 7 dB lower intensity than the others, and participants were instructed to press a button whenever this occurred.
    \item B2021 \cite{boebinger2021music}: This dataset contains fMRI recordings from 20 participants, 10 with musical training (8 female, 2 male; mean age = 23.5 years, SD = 3.3; 11–23 years of training) and 10 without musical training (6 female, 4 male; mean age = 25.8 years, SD = 4.1). The auditory stimuli were identical to those used in NH2015, and the experimental design was similar. The main differences were that each stimulus was repeated 3 rather than 5 times per block, 2 blocks were presented per stimulus (resulting in 6 presentations per session), the experiment was divided into 16 runs, and one stimulus was presented at 12 dB lower intensity than the others to monitor attention.
\end{itemize}

In both datasets, blood-oxygen-level-dependent (BOLD) responses were recorded with fMRI from voxels within brain regions that include the auditory cortex \cite{norman2015distinct,boebinger2021music,tuckute2023many}, which can be seen in \citet{boebinger2021music}. Data were acquired in an orientation parallel to the superior temporal plane, covering the superior temporal gyrus and the superior temporal sulcus. For each session, participant, and stimulus, the first acquisition was discarded. In NH2015, the BOLD response for the remaining four repetitions of each stimulus were averaged, whereas in B2021 a general linear model (GLM) was used instead. The combined signals were converted to percent signal change (PSC) by subtracting and dividing by the voxel’s response to silence, and subsequently averaged over time to yield a single scalar response.

For voxel selection, as done by \citet{tuckute2023many}, we retained those that exhibited significantly greater responses for the stimuli than for silence (t-test, $p<0.001$) and that responded consistently across sessions. Consistency was quantified for each participant as:

\begin{equation}
    r = 1 - \frac{||v_{12}-\frac{v_3 \cdot v_{12}}{||v_3||^2}v_3||_2}{||v_{12}||_2} 
\end{equation}

where, for NH2015, $v_{12}$ denotes a voxel’s response to the 165 stimuli averaged over the first two sessions, and $v_3$ its response in the third session. For B2021, $v_{12}$ corresponds to the responses estimated from the first three repetitions of each stimulus in runs 1–24, and $v_3$ from the last three repetitions in runs 25–48. Voxels with $r>0.3$ were included. This selection process is done separately for each subject. This procedure yielded an average of 961.75 voxels per participant (range 637–1221) in NH2015, and 1340 (range 1020–1828) in B2021. Finally, the responses for the selected voxels were averaged across sessions to get a single value per selected voxel per subject.
 
\subsection{Audio models} \label{sec:models}

We analyzed 36 different models, spanning a wide range of model sizes, objectives and audio domains.

\begin{itemize}
    \item \textbf{EnCodecMAE} \cite{pepino25_interspeech} is a Masked Autoencoder transformer pretrained with a masked audio modelling pretext task using different datasets. A fraction (50\%) of the input frames are masked and the model is trained to predict discrete targets corresponding to the masked segments using the unmasked ones. In a first iteration, discrete labels are obtained from EnCodec \cite{defossez2022highfi}, a neural audio codec, and in a second iteration, the outputs from the first iteration model are quantized with K-Means and used as targets.

    We experimented with 10 different variants:
    \begin{enumerate}
        \item \textbf{ECMAE (B)}: first iteration base model with 10 transformer layers, 86.6 million parameters and 768-dimensional activations, using melspectrograms as inputs and pretrained with a mixture of Audioset (AS), LibriLight (LL) and Free Music Archive (FMA). Audioset consists of around 5000 hours of audio from YouTube videos containing a variety of acoustic events, speech and music. LibriLight is a large scale dataset containing speech recordings from audiobooks. EnCodecMAE uses a subset with 6000 hours of read speech recordings. Free Music Archive consists of 800 hours of music.
        \item \textbf{ECMAE (AS)}: same as 1) but pretrained only with AS.
        \item \textbf{ECMAE (LL)}: same as 1) but pretrained only with LL.
        \item \textbf{ECMAE (FMA)}: same as 1) but pretrained only with FMA.
        \item \textbf{ECMAE (S)}: small version of 1) with only 5 layers and around 43 million parameters.
        \item \textbf{ECMAE (L)}: large version of 1) but with 20 layers and 1024-dimensional activations, resulting in around 261 million parameters.
        \item \textbf{ECMAE (EC)}: same as 1) but using EnCodec features as input instead of melspectrograms.
        \item \textbf{ECMAE (Spec)}: same as 1) but using linear spectrograms as input instead of melspectrograms.
        \item \textbf{ECMAE (It 2)}: model resulting from the second training iteration, starting from 1).
        \item \textbf{ECMAE (L + It 3)}: version of 6) trained for 150k extra steps using quantized outputs of 9) as targets, hence being a third training iteration starting with 1).
        
   \end{enumerate}

    For the small and base models, we extracted activations from the output of every transformer layer, while for the large models, we extracted activations from every odd layer (1, 3, 5,..., 19) and the last layer (20). 
    \item \textbf{BEATs} \cite{chen2023beats} consists of a 12-layer Vision Transformer, pretrained with Audioset\cite{audioset} in a self-supervised way with the pretext task of masked audio modelling (MAM). A fraction (75\%) of the melspectrogram input patches are masked and the model is trained to predict discrete labels corresponding to the masked patches using the unmasked ones. In the first iteration, the discrete labels are generated with a random projection tokenizer as in \cite{chiu2022self}. For the second and third iterations, a tokenizer is trained to discretize the outputs of the previous BEATs iteration, and used to generate the discrete  pretraining targets. 

    We experimented with 4 checkpoints from this model: the final one for each of the 3 iterations \textbf{(BEATs it 1, 2 and 3)}, and \textbf{BEATs (FT)} where the checkpoint from the third iteration is finetuned with Audioset for acoustic event detection. In all cases, we extracted 13 sets of activations: the output of each of the 12 transformer layers and the input of the first transformer layer. All the activations have 768 dimensions.
    
    \item \textbf{Dasheng} \cite{dinkel2024dasheng} is a Masked Autoencoder transformer pretrained with a masked audio modeling task with 272K hours of audio from diverse domains. Instead of using discrete labels as targets, the mean squared error between the reconstructed melspectrogram and the unmasked one is used as a loss. 
    
    We experimented with 4 checkpoints from this model: a base model, \textbf{Dasheng (B)}, with 86M parameters, 12 layers and 768-dimensional activations; a 0.6B parameters model, \textbf{Dasheng (XL)}, with 32 layers and 1024-dimensional activations, a 1.2B parameters model, \textbf{Dasheng (XXL)}, with 40 layers and 1536-dimensional activations; and \textbf{Dasheng (FT)}, a base model fine-tuned in Audioset for acoustic event detection. For the base models, we extracted embeddings from the output of every transformer layer, while for the 0.6B model we extracted from every 3 layers (1, 4, 7,..., 31) and the last layer (32), and for the 1.2B model we extracted from every 4 layers (1, 5, 9, 13,..., 37) and the last layer (40).
    
    \item \textbf{CochDNN} \cite{tuckute2023many} takes as input cochleagrams computed using a bank of 211 band-pass filters designed to approximate the response of the human ear. Envelope extraction is performed via a Hilbert transform, followed by a compression stage simulating that of the basilar membrane. The resulting envelopes are then low-pass filtered and downsampled to 200 Hz. Two convolutional architectures, one with 9 layers (\textbf{CNN9}), and one based in ResNet50 \textbf{CRN50}, were trained on a dataset in which each example consisted of a spoken word embedded in background noise sampled from AudioSet. This setup enabled training on three tasks using the same data: word recognition (\textbf{Word}), speaker identification (\textbf{Spk}), and acoustic event detection based on the background noise (\textbf{AS}). In addition, the authors trained models jointly on the three tasks in a multitask setting (\textbf{MT}). Combining the two architectures with the four objectives resulted in a total of 8 models.
    
    Following \citet{tuckute2023many}, for CNN9 we extracted features from the outputs of the ReLU activations and pooling layers at each convolutional block, leading to 9 sets of  activations with shapes ranging from 2304 to 9216 dimensions, and for CRN50 we extracted features from the outputs of the first convolutional layer after applying ReLU, the outputs of each ResNet block, and the output of the last average pooling layer, leading to 7 sets of activations with dimensions ranging from 2048 to 14336.
    
    \item \textbf{AST} \cite{gong2021ast} is a model that applies a vision transformer to melspectrogram inputs and is trained for acoustic event detection in AudioSet. It is composed of an initial patching and projection layer, which takes the melspectrogram input, splits it into a sequence of 16x16 patches, and projects each of them into 768D vectors. Twelve transformer blocks process this sequence of projected patches and a final linear layer performs the classification into 527 different classes. We extracted embeddings from the output of the patching layer, the 12 transformer blocks, and the final classification layer, leading to 14 sets of activations with 768 dimensions, except for the last layer which has 527 dimensions.
    
    \item \textbf{VGGish} \cite{hershey2017cnn} is an audio classification model that follows the VGG \cite{simonyan2015very} convolutional neural network architecture. The model was trained on the YouTube-100M corpus (70M training videos, 5.24 million hours with 30,871 labels) to predict the video-level labels based on audio information using a cross-entropy loss function. Following \citet{tuckute2023many}, we extracted representations from the ReLU activation after each convolutional block and dense layer, and from the max-pooling layers. This led to 13 sets of activations having between 128 and 4096 dimensions.
    
    \item \textbf{DCASE 2020} \cite{dcase2020} is a recurrent neural network trained for audio captioning, where the model accepts audio as input and outputs a description of it. We used a model trained in the Clotho dataset \cite{drossos2020clotho}, which consists of pairs of audios and captions, and was released by the authors. The input to the model is a log-melspectrogram with 64 mel filters, a 23 ms window size, and a 11.5 ms hop size. The model consists of an encoder with 3 bidirectional Gated Recurrent Units (GRU), and a decoder with 1 GRU and an output layer with 4367 neurons corresponding to each unique word in the captions. We extracted activations from the output of every GRU layer and the final output layer, leading to 5 sets of activations with dimensionalities of 512 for the BiGRUs in the encoder, 256 for the decoder GRU and 4367 for the output layer.
    \item \textbf{DeepSpeech 2} \cite{amodei2016deep} is a recurrent neural network trained for automatic speech recognition (ASR). The input to the model is a log-spectrogram, which is processed by two 2D convolutional layers followed by 5 BiLSTM layers and a final fully-connected layer that outputs logits for the 29 classes corresponding to English characters, space, blank and apostrophe. We extracted activations from the 2 convolutional layers and the 5 BiLSTM layers, leading to 7 sets of activations with dimensions ranging from 1312 to 2592.
    \item \textbf{MetricGAN} \cite{fu21_interspeech} is a generative adversarial network (GAN) trained for speech enhancement. We used a model trained on the Voicebank-DEMAND dataset \cite{veaux2013voice}, which consists of pairs of clean high-quality speech and noise recordings, and is used to train speech enhancement models. The generator consists of 2 BiLSTM layers followed by 2 fully connected layers, while the discriminator is a CNN. The input to the model is a linear spectrogram with a window size of 32 ms and a hop size of 16 ms. We extracted activations from the cell state of the 2 BiLSTM layers and from the outputs of the 2 linear layers, leading to 4 sets of activations with 400 dimensions for the BiLSTM layers, and 300 and 257 dimensions for the linear layers.
    
    \item \textbf{Wav2Vec 2.0} \cite{baevski2020wav2vec} is a self-supervised speech model pretrained to reconstruct masked segments from speech inputs, and finetuned for ASR. It consists of an initial CNN encoder followed by 12 transformer blocks. Activations were extracted from the CNN encoder output and each of the 12 transformer blocks outputs resulting in 13 sets of activations with 768 dimensions each. We used the base version pretrained and finetuned on the LibriSpeech corpus (960 hours) \cite{panayotov2015librispeech}.

    \item \textbf{Sepformer} \cite{subakan2021attention} is a model trained for speech separation with the WHAMR! \cite{Maciejewski2020WHAMR} dataset, and the model estimates optimal masks to separate the speakers from the mixture. The architecture consists of an initial CNN encoder layer, followed by 32 dual-path transformer blocks, which consist of Intra-Transformers modeling short-term temporal dependencies, and Inter-Transformers modeling longer term dependencies. We extracted activations from the output of the initial encoder layer and every transformer layer, leading to 33 sets of activations with 256 dimensions.
    
    \item \textbf{S2T} \cite{wang2020fairseq} is an encoder-decoder model trained for speech-to-text tasks such as ASR and speech to text translation. The model consists of 2 convolutional layers followed by 12 transformer blocks. We used the checkpoint trained for ASR using Librispeech (960h). We extracted activations from the output of the CNN encoder and each transformer layer, leading to 13 sets of activations, each with 1024 dimensions.
    \item \textbf{ZeroSpeech} \cite{van2020vector} consists of a Vector-quantized Variational Autoencoder (VQ-VAE), trained for speech reconstruction, and can be used to generate speech in a target speaker's voice. The encoder is a CNN, while the decoder is a RNN. We only used the encoder, which takes log-melspectrograms as inputs and is followed by 5 1D convolutional layers. We extracted activations from the outputs of every convolutional layer (after ReLU is applied), leading to 5 sets of activations, each with 768 dimensions.
    
    \item \textbf{Spectro-temporal model} \cite{chi2005multiresolution} consists of a linear filter bank tuned to spectrotemporal modulations at different frequencies, spectral scales and temporal rates. Spectrotemporal modulation filters were implemented as 2D convolutions with zero padding in frequency (800 samples) and time (211 samples). Filters targeted spectral modulation frequencies of 0.0625–2 cycles/erb and temporal modulation frequencies of 0.5–64 Hz, including both upward and downward sweeps, yielding 96 filters. To capture purely spectral and purely temporal modulations, we added 6 and 8 filters, respectively, for a total of 110. Filter outputs were squared and averaged over time within each frequency channel.
\end{itemize}

\subsection{Voxel response regression}
\label{sec:methods-regresion}
For each subject $s$ and each voxel $v$, we trained an L2-regularized (ridge) linear regressor to predict its response using the audio representations $X_{m,l}$ from layer $l$ of a model $m$ as input. The output of the regressor is a vector of predictions $\hat{Y}_{m,l,s,v}$, which is compared with the target $Y_{s,v}$ using the Pearson determination coefficient and resulting in a matrix $R^2_{m,l,s,v}$. If the Pearson correlation coefficient was less than 0, we set it to 0, as in \citet{tuckute2023many}. For each model $m$, subject $s$, and voxel $v$, we performed a hyperparameter search to select the regularization strength ($\alpha$) and the layer ($l$) to use for prediction. We explored the following $\alpha$ values: $[0.01, 0.05, 0.1, 0.5, 1.0, 5.0, 10.0, 50.0]$. If the best value of alpha was at the edges ($0.01$ or $50$), we divided or multiplied the best $\alpha$ by 2 until the performance stopped improving, or until $\alpha$ was out of the interval $[10^{-49}, 10^{50}]$.

We divided the 165 stimuli into 5 equal-size splits. The splits were designed so that they were balanced across the 11 different stimuli categories (mechanical sounds, human vocalizations, human non-vocal sounds, english speech, non-english speech, environmental sound, music, animal vocalizations, animal non-vocal sounds, song and nature sounds). We performed nested cross-validation to select the hyperparameters ($\alpha$ and model layer). That is, for each split, we used the remaining 4 splits to select the hyperparameters using an inner loop of cross-validation. Then, we trained the model in those 4 splits, using the hyperparameter set selected in the inner loop, and predicted the voxel response for the remaining split. This process was repeated until there were predictions for every split. This way, we obtained the Pearson determination coefficient $R^2_{m,s,v}$ between the voxel response predicted for the 165 stimuli and the fMRI measurements at voxel $v$ of subject $s$, for audio model $m$. Finally, in order to get a summary metric for the alignment between each model and all brain measurements, we took the median of all the $R^2_{m,s,v}$ for each subject, leading to $R^2_{m,s}$, and then took the mean across subjects obtaining the final summary metric $R^2_m$.

\subsection{Component regression} \label{sec:brain-components}

In addition to performing regression directly on voxels, for the NH2015, we also regressed onto independent components obtained in a previous study 
\cite{norman2015distinct}. The decomposition is done by first averaging voxel responses over all subjects and then decomposing the resulting matrix of dimension given by the number of stimuli ($N=165$) times the number of voxels. The authors identified six independent components that together explained 80\% of the voxel variance in the NH2015 dataset. Moreover, the authors were able to characterize the stimulus selectivity of each component:
\begin{itemize}
    \item \textbf{Component 1 (LF)} assigned higher weights to voxels responding to low-frequency pure tones in a tonotopy mapping experiment.
    \item \textbf{Component 2 (HF)} assigned higher weights to voxels responding to high-frequency pure tones.
    \item \textbf{Component 3 (Broadband)} was selective for stimuli with broadband spectra and rapid temporal modulations, exhibiting vertical patterns in a spectrogram.
    \item \textbf{Component 4 (Pitch)} was selective for tonal stimuli, exhibiting horizontal lines in a spectrogram.
    \item \textbf{Component 5 (Speech)} showed strong selectivity for speech, with speech stimuli eliciting the highest responses, followed by sung music and vocalizations.
    \item \textbf{Component 6 (Music)} showed selective responses to music. 
\end{itemize}

For consistency, we report the components in the same order as the \citet{norman2015distinct} and \citet{tuckute2023many} studies. 
The reported metric is the Pearson coefficient of determination $R^2$ between the predicted component values and the ones measured from fMRI data.

\subsection{Representation similarity analysis} \label{sec:methods-rsa}

Given a matrix $M$, a representation dissimilarity matrix (RDM) $D$ can be obtained as

\begin{equation}
    D_{ij} = 1 - r(M_i, M_j)
\end{equation}

with $r(M_i, M_j)$ being a similarity measure between the rows $i$ and $j$ of the $M$ matrix. We used the Pearson correlation coefficient as a similarity measure, and obtained dissimilarity matrices $D^a_l$ for the audio model responses $M^a_l$ at each layer $l$, and dissimilarity matrices $D^b_s$ for the voxel responses $M^b_s$ of each subject $s$. The  matrices $D^a_l$ and $D^b_s$ have the same shape $N \times N$ with $N$ being the number of audio stimuli. This consistency in the shape and correspondance of each coefficient to the same pair of stimuli enables the comparison of RDMs coming from representations of different shapes. The upper triangular matrix elements of the RDM are extracted and flattened, and Spearman correlation is measured between the resulting vectors, yielding a single scalar metric.

The above approach results in a coefficient $\rho_{m,l,s}$ for every subject $s$ and every layer $l$ of the audio model $m$, which reflects the similarity between the representations obtained from the fMRI of subject $s$ and the representations obtained from the layer $l$ of audio model $m$. In order to summarize $\rho_{m,l,s}$ into a single scalar $\rho_m$ per model $m$, we computed the average of $\rho_{m,l,s}$ across subjects, leading to a Spearman correlation coefficient for each layer $\rho_{m,l}$. We finally computed the maximum value across layers as summary metric. We also explored using cross-validation for layer selection (see Supplementary material), but differences were minimal and we decided to use the simpler method.

In order to obtain RSA values for the primary and posterior regions, 
we constructed the fMRI RDM matrices $D^b_s$ using only the subset of voxels corresponding to the anatomical region of interest.

\subsection{Audio models performance measurement}

In order to measure how good the model representations are for solving audio-related tasks, we evaluated the models in a subset of the HEAREval benchmark \cite{turian2022hear} tasks:

\begin{itemize}
    \item \textbf{Music note classification}: The NSynth dataset \cite{engel2017neural} is used, which contains 4-seconds long isolated music notes from different musical instruments. The goal is to predict which note was played among 88 options. 
    \item \textbf{Music genre classification}: The GTZAN dataset \cite{tzanetakis2002musical} is used, which contains 30 seconds segments of 1000 different songs, categorized in 10 music genres: blues, classical, country, disco, hiphop, jazz, metal, pop, reggae and rock.
    \item \textbf{Speech commands classification}: The goal is to classify which of 10 possible commands were spoken in a speech segment. Two additional categories are 'noise' and 'other'. The Google Speech Commands Dataset is used.
    \item \textbf{Speech emotion recognition}: The CREMA-D dataset \cite{cao2014crema} is used, which contains 12 different sentences said by 91 actors with one of the following emotions: anger, disgust, fear, happiness, sadness and neutral, and different emotional intensities. The task consists in classifying the emotion from the speech signal.
    \item \textbf{Acoustic event detection}: The FSD-50K dataset  \cite{fonseca2021fsd50k} is used, which consists of 100 hours of audios annotated with acoustic events, which can belong to 200 different categories. The task consists in determining which acoustic events happened in an audio segment. As multiple events can occur in a single segment, this is a multilabel classification problem.
    \item \textbf{Environmental sound classification}: The ESC-50 dataset \cite{piczak2015esc} is used, which consists of 2000 5-seconds long recordings of isolated environmental sounds, which can belong to 50 different categories. Some examples of the categories are: rain, dog, baby crying, door slam and helicopter.
\end{itemize}

We used accuracy as the evaluation metric, except for acoustic event detection, where we used mean average precision (mAP) instead. Together, these tasks assess model performance in diverse domains (speech, music and environmental sounds) using well established datasets.

For each task, we train a downstream classifier that takes the audio representations as input and predicts the corresponding labels. To be consistent with the regression and RSA analysis, we used the representations from the same layers to train the downstream model. Given a set of $L$ representations $\{X_{m,1}, X_{m,2},..., X_{m,L}\}$ from the audio model $m$, we learn a set of $L$ weights $\{\alpha_1, \alpha_2,..., \alpha_L\}$ and obtain a single representation $X_m$ by performing a weighted average of the representations:

\begin{equation}
    X_m = \frac{\sum_{i=1}^L |\alpha_i|X_{m,i}}{\sum_{i=1}^L |\alpha_i|}
\end{equation}

When the representations $X_{m,i}$ have different dimensionalities, we perform Principal Component Analysis (PCA) for each representation and replace $X_{m,i}$ by its first $D_{\min}$ components, where $D_{\min}$ is the smallest dimensionality among the representations. The PCA components are estimated using the downstream task training data.

A multilayer perceptron taking $X_m$ as input is trained jointly with the weights $\alpha_i$ to predict the targets for the task. A small hyperparameter search is performed. Table \ref{tab:hyp-hear} shows the learning rate, number of hidden layers, and initialization explored, leading to 16 hyperparameter combinations. The search is done using the validation set, when the datasets have train-validation-test splits, and is done in the first fold when the datasets have $K$ cross-validation folds (by using $K-2$ folds for training, 1 for validation and determining early stopping and 1 for test and determining hyperparameter performance).

\begin{table}[h!]
\centering
\begin{tabular}{ll}
\hline
\textbf{Searched hyperparameters} & \textbf{Values} \\
\hline
Hidden layers & \{1, 2\} \\
Learning rate & \{3.2$\times$10$^{-3}$, 1$\times$10$^{-3}$, 3.2$\times$10$^{-4}$, 1$\times$10$^{-4}$\} \\
Initialization & \{Xavier Uniform, Xavier Normal\} \\
\hline
\textbf{Fixed hyperparameters} & \\
\hline
Dropout & 0.1 \\
Normalization & BatchNorm \\
Hidden activation & ReLU \\
Hidden neurons & 1024 \\
\hline
\end{tabular}
\caption{Downstream model hyperparameters, following HEAREval benchmark methodology.}
\label{tab:hyp-hear}
\end{table}

Finally, to obtain an overall measure of model performance across tasks, we z-scored metrics within each task and averaged across tasks to compute a single global score.

\subsection{Code and Data Availability}
The source code and results from this work can be found at \url{https://github.com/mrpep/braindnn}. Brain measurements can be obtained from \url{https://mcdermottlab.mit.edu//tuckute_feather_2023/data.tar} \cite{norman2015distinct,boebinger2021music,tuckute2023many}, and datasets for downstream evaluation can be downloaded from \url{https://zenodo.org/records/5887964} \cite{turian2022hear}.

\subsection{Acknowledgements}
The authors were supported by the Consejo Nacional de Investigaciones Cientificas y Tecnicas (CONICET, Argentina) and the Universidad de Buenos Aires (UBA, Argentina).

\bibliography{sn-bibliography}

@article{tzanetakis2002musical,
  title={Musical genre classification of audio signals},
  author={Tzanetakis, George and Cook, Perry},
  journal={IEEE Transactions on speech and audio processing},
  volume={10},
  number={5},
  pages={293--302},
  year={2002},
  publisher={IEEE}
}

@article{baevski2020wav2vec,
  title={wav2vec 2.0: A framework for self-supervised learning of speech representations},
  author={Baevski, Alexei and Zhou, Yuhao and Mohamed, Abdelrahman and Auli, Michael},
  journal={Advances in neural information processing systems},
  volume={33},
  pages={12449--12460},
  year={2020}
}

@inproceedings{hershey2017cnn,
  title={CNN architectures for large-scale audio classification},
  author={Hershey, Shawn and Chaudhuri, Sourish and Ellis, Daniel PW and Gemmeke, Jort F and Jansen, Aren and Moore, R Channing and Plakal, Manoj and Platt, Devin and Saurous, Rif A and Seybold, Bryan and others},
  booktitle={2017 ieee international conference on acoustics, speech and signal processing (icassp)},
  pages={131--135},
  year={2017},
  organization={IEEE}
}

@article{gong2021ast,
  title={Ast: Audio spectrogram transformer},
  author={Gong, Yuan and Chung, Yu-An and Glass, James},
  journal={arXiv preprint arXiv:2104.01778},
  year={2021}
}

@INPROCEEDINGS{audioset,
  author={Gemmeke, Jort F. and Ellis, Daniel P. W. and Freedman, Dylan and Jansen, Aren and Lawrence, Wade and Moore, R. Channing and Plakal, Manoj and Ritter, Marvin},
  booktitle={2017 IEEE International Conference on Acoustics, Speech and Signal Processing (ICASSP)}, 
  title={Audio Set: An ontology and human-labeled dataset for audio events}, 
  year={2017},
  volume={},
  number={},
  pages={776-780},
  doi={10.1109/ICASSP.2017.7952261}}

@inproceedings{chiu2022self,
  title={Self-supervised learning with random-projection quantizer for speech recognition},
  author={Chiu, Chung-Cheng and Qin, James and Zhang, Yu and Yu, Jiahui and Wu, Yonghui},
  booktitle={International Conference on Machine Learning},
  pages={3915--3924},
  year={2022},
  organization={PMLR}
}

@inproceedings{chen2023beats,
  title={BEATs: audio pre-training with acoustic tokenizers},
  author={Chen, Sanyuan and Wu, Yu and Wang, Chengyi and Liu, Shujie and Tompkins, Daniel and Chen, Zhuo and Che, Wanxiang and Yu, Xiangzhan and Wei, Furu},
  booktitle={Proceedings of the 40th International Conference on Machine Learning},
  pages={5178--5193},
  year={2023}
}

@inproceedings{engel2017neural,
  title={Neural audio synthesis of musical notes with wavenet autoencoders},
  author={Engel, Jesse and Resnick, Cinjon and Roberts, Adam and Dieleman, Sander and Norouzi, Mohammad and Eck, Douglas and Simonyan, Karen},
  booktitle={International Conference on Machine Learning},
  pages={1068--1077},
  year={2017},
  organization={PMLR}
}

@article{defossez2022highfi,
  title={High Fidelity Neural Audio Compression},
  author={Défossez, Alexandre and Copet, Jade and Synnaeve, Gabriel and Adi, Yossi},
  journal={arXiv preprint arXiv:2210.13438},
  year={2022}
}

@article{tuckute2023many,
  title={Many but not all deep neural network audio models capture brain responses and exhibit correspondence between model stages and brain regions},
  author={Tuckute, Greta and Feather, Jenelle and Boebinger, Dana and McDermott, Josh H},
  journal={Plos Biology},
  volume={21},
  number={12},
  pages={e3002366},
  year={2023},
  publisher={Public Library of Science San Francisco, CA USA}
}

@article{huh2024platonic,
  title={The platonic representation hypothesis},
  author={Huh, Minyoung and Cheung, Brian and Wang, Tongzhou and Isola, Phillip},
  journal={arXiv preprint arXiv:2405.07987},
  year={2024}
}

@article{chi2005multiresolution,
  title={Multiresolution spectrotemporal analysis of complex sounds},
  author={Chi, Taishih and Ru, Powen and Shamma, Shihab A},
  journal={The Journal of the Acoustical Society of America},
  volume={118},
  number={2},
  pages={887--906},
  year={2005},
  publisher={AIP Publishing}
}

@article{gucclu2016brains,
  title={Brains on beats},
  author={G{\"u}{\c{c}}l{\"u}, Umut and Thielen, Jordy and Hanke, Michael and Van Gerven, Marcel},
  journal={Advances in Neural Information Processing Systems},
  volume={29},
  year={2016}
}

@article{khatami2020spiking,
  title={Spiking network optimized for word recognition in noise predicts auditory system hierarchy},
  author={Khatami, Fatemeh and Escab{\'\i}, Monty A},
  journal={PLOS Computational Biology},
  volume={16},
  number={6},
  pages={e1007558},
  year={2020},
  publisher={Public Library of Science San Francisco, CA USA}
}

@article{millet2021inductive,
  title={Inductive biases, pretraining and fine-tuning jointly account for brain responses to speech},
  author={Millet, Juliette and King, Jean-Remi},
  journal={arXiv preprint arXiv:2103.01032},
  year={2021}
}

@inproceedings{vaidya2022self,
  title={Self-Supervised Models of Audio Effectively Explain Human Cortical Responses to Speech},
  author={Vaidya, Aditya R and Jain, Shailee and Huth, Alexander},
  booktitle={International Conference on Machine Learning},
  pages={21927--21944},
  year={2022},
  organization={PMLR}
}

@article{norman2015distinct,
  title={Distinct cortical pathways for music and speech revealed by hypothesis-free voxel decomposition},
  author={Norman-Haignere, Sam and Kanwisher, Nancy G and McDermott, Josh H},
  journal={neuron},
  volume={88},
  number={6},
  pages={1281--1296},
  year={2015},
  publisher={Elsevier}
}

@inproceedings{dinkel2024dasheng,
  title={Scaling up masked audio encoder learning for general audio classification},
  author={Dinkel, Heinrich and Yan, Zhiyong and Wang, Yongqing and Zhang, Junbo and Wang, Yujun and Wang, Bin},
  booktitle={Interspeech 2024},
  year={2024}
}

@article{fonseca2021fsd50k,
   title={Fsd50k: an open dataset of human-labeled sound events},
   author={Fonseca, Eduardo and Favory, Xavier and Pons, Jordi and Font, Frederic and Serra, Xavier},
   journal={IEEE/ACM Transactions on Audio, Speech, and Language Processing},
   volume={30},
   pages={829--852},
   year={2021},
   publisher={IEEE}
 }

@inproceedings{turian2022hear,
   title={Hear: Holistic evaluation of audio representations},
   author={Turian, Joseph and Shier, Jordie and Khan, Humair Raj and Raj, Bhiksha and Schuller, Bj{\"o}rn W and Steinmetz, Christian J and Malloy, Colin and Tzanetakis, George and Velarde, Gissel and McNally, Kirk and others},
   booktitle={NeurIPS 2021 Competitions and Demonstrations Track},
   pages={125--145},
   year={2022},
   organization={PMLR}
 }

@article{cao2014crema,
   title={Crema-d: Crowd-sourced emotional multimodal actors dataset},
   author={Cao, Houwei and Cooper, David G and Keutmann, Michael K and Gur, Ruben C and Nenkova, Ani and Verma, Ragini},
   journal={IEEE transactions on affective computing},
   volume={5},
   number={4},
   pages={377--390},
   year={2014},
   publisher={IEEE}
 }

@inproceedings{piczak2015esc,
   title={ESC: Dataset for environmental sound classification},
   author={Piczak, Karol J},
   booktitle={Proceedings of the 23rd ACM international conference on Multimedia},
   pages={1015--1018},
   year={2015}
 }

@inproceedings{panayotov2015librispeech,
  title={Librispeech: an asr corpus based on public domain audio books},
  author={Panayotov, Vassil and Chen, Guoguo and Povey, Daniel and Khudanpur, Sanjeev},
  booktitle={2015 IEEE international conference on acoustics, speech and signal processing (ICASSP)},
  pages={5206--5210},
  year={2015},
  organization={IEEE}
}

@inproceedings{lee2021acav100m,
  title={Acav100m: Automatic curation of large-scale datasets for audio-visual video representation learning},
  author={Lee, Sangho and Chung, Jiwan and Yu, Youngjae and Kim, Gunhee and Breuel, Thomas and Chechik, Gal and Song, Yale},
  booktitle={Proceedings of the IEEE/CVF International Conference on Computer Vision},
  pages={10274--10284},
  year={2021}
}

@inproceedings{awinstruction,
  title={Instruction-tuning Aligns LLMs to the Human Brain},
  author={Aw, Khai Loong and Montariol, Syrielle and AlKhamissi, Badr and Schrimpf, Martin and Bosselut, Antoine},
  booktitle={First Conference on Language Modeling}
}

@article{boebinger2021music,
  title={Music-selective neural populations arise without musical training},
  author={Boebinger, Dana and Norman-Haignere, Sam V and McDermott, Josh H and Kanwisher, Nancy},
  journal={Journal of Neurophysiology},
  volume={125},
  number={6},
  pages={2237--2263},
  year={2021},
  publisher={American Physiological Society Rockville, MD}
}

@article{doerig2023neuroconnectionist,
  title={The neuroconnectionist research programme},
  author={Doerig, Adrien and Sommers, Rowan P and Seeliger, Katja and Richards, Blake and Ismael, Jenann and Lindsay, Grace W and Kording, Konrad P and Konkle, Talia and Van Gerven, Marcel AJ and Kriegeskorte, Nikolaus and others},
  journal={Nature Reviews Neuroscience},
  volume={24},
  number={7},
  pages={431--450},
  year={2023},
  publisher={Nature Publishing Group UK London}
}

@inproceedings{amodei2016deep,
  title={Deep speech 2: End-to-end speech recognition in english and mandarin},
  author={Amodei, Dario and Ananthanarayanan, Sundaram and Anubhai, Rishita and Bai, Jingliang and Battenberg, Eric and Case, Carl and Casper, Jared and Catanzaro, Bryan and Cheng, Qiang and Chen, Guoliang and others},
  booktitle={International conference on machine learning},
  pages={173--182},
  year={2016},
  organization={PMLR}
}

@inproceedings{wang2020fairseq,
  title={Fairseq S2T: Fast Speech-to-Text Modeling with Fairseq},
  author={Wang, Changhan and Tang, Yun and Ma, Xutai and Wu, Anne and Okhonko, Dmytro and Pino, Juan},
  booktitle={Proceedings of the 1st Conference of the Asia-Pacific Chapter of the Association for Computational Linguistics and the 10th International Joint Conference on Natural Language Processing: System Demonstrations},
  pages={33--39},
  year={2020}
}

@inproceedings{subakan2021attention,
  title={Attention is all you need in speech separation},
  author={Subakan, Cem and Ravanelli, Mirco and Cornell, Samuele and Bronzi, Mirko and Zhong, Jianyuan},
  booktitle={ICASSP 2021-2021 IEEE International Conference on Acoustics, Speech and Signal Processing (ICASSP)},
  pages={21--25},
  year={2021},
  organization={IEEE}
}

@article{van2020vector,
  title={Vector-Quantized Neural Networks for Acoustic Unit Discovery in the ZeroSpeech 2020 Challenge},
  author={van Niekerk, Benjamin and Nortje, Leanne and Kamper, Herman},
  journal={Interspeech 2020},
  pages={4836--4840},
  year={2020},
  publisher={ISCA}
}

@article{moussa2024improving,
  title={Improving semantic understanding in speech language models via brain-tuning},
  author={Moussa, Omer and Klakow, Dietrich and Toneva, Mariya},
  journal={arXiv preprint arXiv:2410.09230},
  year={2024}
}

@article{freteault2025alignment,
  title={Alignment of auditory artificial networks with massive individual fMRI brain data leads to generalisable improvements in brain encoding and downstream tasks},
  author={Freteault, Maelle and Le Clei, Maximilien and Tetrel, Loic and Bellec, Lune and Farrugia, Nicolas},
  journal={Imaging Neuroscience},
  volume={3},
  pages={imag\_a\_00525},
  year={2025},
  publisher={MIT Press 255 Main Street, 9th Floor, Cambridge, Massachusetts 02142, USA~…}
}

@article{rauch2025can,
  title={Can Masked Autoencoders Also Listen to Birds?},
  author={Rauch, Lukas and Moummad, Ilyass and Heinrich, Ren{\'e} and Joly, Alexis and Sick, Bernhard and Scholz, Christoph},
  journal={arXiv preprint arXiv:2504.12880},
  year={2025}
}

@inproceedings{hagiwara2023aves,
  title={AVES: Animal vocalization encoder based on self-supervision},
  author={Hagiwara, Masato},
  booktitle={ICASSP 2023-2023 IEEE International Conference on Acoustics, Speech and Signal Processing (ICASSP)},
  pages={1--5},
  year={2023},
  organization={IEEE}
}

@article{moummad2024domain,
  title={Domain-Invariant Representation Learning of Bird Sounds},
  author={Moummad, Ilyass and Serizel, Romain and Benetos, Emmanouil and Farrugia, Nicolas},
  journal={arXiv preprint arXiv:2409.08589},
  year={2024}
}

@article{gucclu2015deep,
  title={Deep neural networks reveal a gradient in the complexity of neural representations across the ventral stream},
  author={G{\"u}{\c{c}}l{\"u}, Umut and Van Gerven, Marcel AJ},
  journal={Journal of Neuroscience},
  volume={35},
  number={27},
  pages={10005--10014},
  year={2015},
  publisher={Society for Neuroscience}
}

@article{seeliger2018convolutional,
  title={Convolutional neural network-based encoding and decoding of visual object recognition in space and time},
  author={Seeliger, Katja and Fritsche, Matthias and G{\"u}{\c{c}}l{\"u}, Umut and Schoenmakers, Sanne and Schoffelen, J-M and Bosch, Sander E and Van Gerven, MAJ},
  journal={NeuroImage},
  volume={180},
  pages={253--266},
  year={2018},
  publisher={Elsevier}
}

@article{eickenberg2017seeing,
  title={Seeing it all: Convolutional network layers map the function of the human visual system},
  author={Eickenberg, Michael and Gramfort, Alexandre and Varoquaux, Ga{\"e}l and Thirion, Bertrand},
  journal={NeuroImage},
  volume={152},
  pages={184--194},
  year={2017},
  publisher={Elsevier}
}

@article{yamins2014performance,
  title={Performance-optimized hierarchical models predict neural responses in higher visual cortex},
  author={Yamins, Daniel LK and Hong, Ha and Cadieu, Charles F and Solomon, Ethan A and Seibert, Darren and DiCarlo, James J},
  journal={Proceedings of the national academy of sciences},
  volume={111},
  number={23},
  pages={8619--8624},
  year={2014},
  publisher={National Academy of Sciences}
}

@article{khaligh2014deep,
  title={Deep supervised, but not unsupervised, models may explain IT cortical representation},
  author={Khaligh-Razavi, Seyed-Mahdi and Kriegeskorte, Nikolaus},
  journal={PLoS computational biology},
  volume={10},
  number={11},
  pages={e1003915},
  year={2014},
  publisher={Public Library of Science San Francisco, USA}
}

@article{cadena2019deep,
  title={Deep convolutional models improve predictions of macaque V1 responses to natural images},
  author={Cadena, Santiago A and Denfield, George H and Walker, Edgar Y and Gatys, Leon A and Tolias, Andreas S and Bethge, Matthias and Ecker, Alexander S},
  journal={PLoS computational biology},
  volume={15},
  number={4},
  pages={e1006897},
  year={2019},
  publisher={Public Library of Science San Francisco, CA USA}
}

@article{goldstein2022correspondence,
  title={Correspondence between the layered structure of deep language models and temporal structure of natural language processing in the human brain},
  author={Goldstein, Ariel and Ham, Eric and Nastase, Samuel A and Zada, Zaid and Grinstein-Dabus, Avigail and Aubrey, Bobbi and Schain, Mariano and Gazula, Harshvardhan and Feder, Amir and Doyle, Werner and others},
  journal={BioRxiv},
  pages={2022--07},
  year={2022},
  publisher={Cold Spring Harbor Laboratory}
}

@article{kumar2024shared,
  title={Shared functional specialization in transformer-based language models and the human brain},
  author={Kumar, Sreejan and Sumers, Theodore R and Yamakoshi, Takateru and Goldstein, Ariel and Hasson, Uri and Norman, Kenneth A and Griffiths, Thomas L and Hawkins, Robert D and Nastase, Samuel A},
  journal={Nature communications},
  volume={15},
  number={1},
  pages={5523},
  year={2024},
  publisher={Nature Publishing Group UK London}
}

@article{Schrimpf2020TheNAA,
  title={The neural architecture of language: Integrative modeling converges on predictive processing},
  author={Martin Schrimpf and I. Blank and Greta Tuckute and Carina Kauf and Eghbal A. Hosseini and N. Kanwisher and J. Tenenbaum and Evelina Fedorenko},
  journal={Proceedings of the National Academy of Sciences},
  year={2020},
  volume={118},
  url={https://doi.org/10.1101/2020.06.26.174482}
}

@inproceedings{pepino25_interspeech,
  title     = {EnCodecMAE: leveraging neural codecs for universal audio representation learning},
  author    = {Leonardo Pepino and Pablo Riera and Luciana Ferrer},
  year      = {2025},
  booktitle = {Interspeech 2025},
  pages     = {3519--3523},
  doi       = {10.21437/Interspeech.2025-506},
  issn      = {2958-1796},
}

@inproceedings{simonyan2015very,
  title={Very deep convolutional networks for large-scale image recognition},
  author={Simonyan, K and Zisserman, A},
  booktitle={3rd International Conference on Learning Representations (ICLR 2015)},
  year={2015},
  organization={Computational and Biological Learning Society}
}

@INPROCEEDINGS{dcase2020,
  author={Drossos, Konstantinos and Adavanne, Sharath and Virtanen, Tuomas},
  booktitle={2017 IEEE Workshop on Applications of Signal Processing to Audio and Acoustics (WASPAA)}, 
  title={Automated audio captioning with recurrent neural networks}, 
  year={2017},
  volume={},
  number={},
  pages={374-378},
  doi={10.1109/WASPAA.2017.8170058}}

@inproceedings{drossos2020clotho,
  title={Clotho: An audio captioning dataset},
  author={Drossos, Konstantinos and Lipping, Samuel and Virtanen, Tuomas},
  booktitle={ICASSP 2020-2020 IEEE International Conference on Acoustics, Speech and Signal Processing (ICASSP)},
  pages={736--740},
  year={2020},
  organization={IEEE}
}

@inproceedings{fu21_interspeech,
  title     = {MetricGAN+: An Improved Version of MetricGAN for Speech Enhancement},
  author    = {Szu-Wei Fu and Cheng Yu and Tsun-An Hsieh and Peter Plantinga and Mirco Ravanelli and Xugang Lu and Yu Tsao},
  year      = {2021},
  booktitle = {Interspeech 2021},
  pages     = {201--205},
  doi       = {10.21437/Interspeech.2021-599},
  issn      = {2958-1796},
}

@inproceedings{veaux2013voice,
  title={The voice bank corpus: Design, collection and data analysis of a large regional accent speech database},
  author={Veaux, Christophe and Yamagishi, Junichi and King, Simon},
  booktitle={2013 international conference oriental COCOSDA held jointly with 2013 conference on Asian spoken language research and evaluation (O-COCOSDA/CASLRE)},
  pages={1--4},
  year={2013},
  organization={IEEE}
}

@inproceedings{Maciejewski2020WHAMR,
    title     = {WHAMR!: Noisy and Reverberant Single-Channel Speech Separation},
    author    = {Maciejewski, Matthew and Wichern, Gordon and Le Roux, Jonathan},
    booktitle = {Proc. IEEE International Conference on Acoustics, Speech and Signal Processing (ICASSP)},
    year      = {2020},
    month     = may
}

@article{mischler2024contextual,
  title={Contextual feature extraction hierarchies converge in large language models and the brain},
  author={Mischler, Gavin and Li, Yinghao Aaron and Bickel, Stephan and Mehta, Ashesh D and Mesgarani, Nima},
  journal={Nature Machine Intelligence},
  volume={6},
  number={12},
  pages={1467--1477},
  year={2024},
  publisher={Nature Publishing Group UK London}
}

@article{Freteault2025AlignmentOA,
  title={Alignment of auditory artificial networks with massive individual fMRI brain data leads to generalisable improvements in brain encoding and downstream tasks},
  author={Ma{\"e}lle Freteault and Lo{\"i}c Tetrel and Maximilien Le Clei and Lune P Bellec and Nicolas Farrugia},
  journal={Imaging Neuroscience},
  year={2025},
  volume={3},
  url={https://api.semanticscholar.org/CorpusID:261661973}
}

@inproceedings{li24l_interspeech,
  title     = {Refining Self-supervised Learnt Speech Representation using Brain Activations},
  author    = {HengYu Li and Kangdi Mei and Zhaoci Liu and Yang Ai and Liping Chen and Jie Zhang and Zhenhua Ling},
  year      = {2024},
  booktitle = {Interspeech 2024},
  pages     = {1480--1484},
  doi       = {10.21437/Interspeech.2024-604},
  issn      = {2958-1796},
}

@article{aw2023instruction,
  title={Instruction-tuning aligns llms to the human brain},
  author={Aw, Khai Loong and Montariol, Syrielle and AlKhamissi, Badr and Schrimpf, Martin and Bosselut, Antoine},
  journal={arXiv preprint arXiv:2312.00575},
  year={2023}
}

@inproceedings{Aw2022TrainingLMA,
  title={Training language models to summarize narratives improves brain alignment},
  author={Khai Loong Aw and Mariya Toneva},
  booktitle={International Conference on Learning Representations},
  year={2022},
  url={https://api.semanticscholar.org/CorpusId:257255248}
}

@inproceedings{huo2025iterative,
  title={Iterative Refinement, Not Training Objective, Makes HuBERT Behave Differently from wav2vec 2.0},
  author={Huo, Robin and Dunbar, Ewan},
  booktitle={Proc. Interspeech 2025},
  pages={261--265},
  year={2025}
}

@article{Hong_2024, title={Scale matters: Large language models with billions (rather than millions) of parameters better match neural representations of natural language}, url={http://dx.doi.org/10.7554/eLife.101204.1}, DOI={10.7554/elife.101204.1}, publisher={eLife Sciences Publications, Ltd}, author={Hong, Zhuoqiao and Wang, Haocheng and Zada, Zaid and Gazula, Harshvardhan and Turner, David and Aubrey, Bobbi and Niekerken, Leonard and Doyle, Werner and Devore, Sasha and Dugan, Patricia and Friedman, Daniel and Devinsky, Orrin and Flinker, Adeen and Hasson, Uri and Nastase, Samuel A and Goldstein, Ariel}, year={2024}, month=oct }

@article{caucheteux2020language,
  title={Language processing in brains and deep neural networks: computational convergence and its limits},
  author={Caucheteux, Charlotte and King, Jean-R{\'e}mi},
  journal={BioRxiv},
  pages={2020--07},
  year={2020},
  publisher={Cold Spring Harbor Laboratory}
}

@article{goldstein2025temporal,
  title={Temporal structure of natural language processing in the human brain corresponds to layered hierarchy of large language models},
  author={Goldstein, Ariel and Ham, Eric and Schain, Mariano and Nastase, Samuel A and Aubrey, Bobbi and Zada, Zaid and Grinstein-Dabush, Avigail and Gazula, Harshvardhan and Feder, Amir and Doyle, Werner and others},
  journal={Nature communications},
  volume={16},
  number={1},
  pages={10529},
  year={2025},
  publisher={Nature Publishing Group UK London}
}

@inproceedings{alkhamissi2025language,
  title={From language to cognition: How llms outgrow the human language network},
  author={AlKhamissi, Badr and Tuckute, Greta and Tang, Yingtian and Binhuraib, Taha Osama A and Bosselut, Antoine and Schrimpf, Martin},
  booktitle={Proceedings of the 2025 Conference on Empirical Methods in Natural Language Processing},
  pages={24332--24350},
  year={2025}
}

@article{wen2018neural,
  title={Neural encoding and decoding with deep learning for dynamic natural vision},
  author={Wen, Haiguang and Shi, Junxing and Zhang, Yizhen and Lu, Kun-Han and Cao, Jiayue and Liu, Zhongming},
  journal={Cerebral cortex},
  volume={28},
  number={12},
  pages={4136--4160},
  year={2018},
  publisher={Oxford University Press}
}

@inproceedings{
  sartzetaki2025one,
  title={One Hundred Neural Networks and Brains Watching Videos: Lessons from Alignment},
  author={Christina Sartzetaki and Gemma Roig and Cees G. M. Snoek and Iris Groen},
  booktitle={The Thirteenth International Conference on Learning Representations},
  year={2025},
  url={https://openreview.net/forum?id=LM4PYXBId5}
}

@article{li2023dissecting,
  title={Dissecting neural computations in the human auditory pathway using deep neural networks for speech},
  author={Li, Yuanning and Anumanchipalli, Gopala K and Mohamed, Abdelrahman and Chen, Peili and Carney, Laurel H and Lu, Junfeng and Wu, Jinsong and Chang, Edward F},
  journal={Nature Neuroscience},
  volume={26},
  number={12},
  pages={2213--2225},
  year={2023},
  publisher={Nature Publishing Group US New York}
}

@article{huang2018connecting,
  title={Connecting deep neural networks to physical, perceptual, and electrophysiological auditory signals},
  author={Huang, Nicholas and Slaney, Malcolm and Elhilali, Mounya},
  journal={Frontiers in neuroscience},
  volume={12},
  pages={532},
  year={2018},
  publisher={Frontiers Media SA}
}

@article{zaiem2025,
title = {Speech self-supervised representations benchmarking: A case for larger probing heads},
journal = {Computer Speech \& Language},
volume = {89},
pages = {101695},
year = {2025},
issn = {0885-2308},
doi = {https://doi.org/10.1016/j.csl.2024.101695},
url = {https://www.sciencedirect.com/science/article/pii/S0885230824000780},
author = {Salah Zaiem and Youcef Kemiche and Titouan Parcollet and Slim Essid and Mirco Ravanelli},
}

@article{li2022petit,
  title={Le Petit Prince multilingual naturalistic fMRI corpus},
  author={Li, Jixing and Bhattasali, Shohini and Zhang, Shulin and Franzluebbers, Berta and Luh, Wen-Ming and Spreng, R Nathan and Brennan, Jonathan R and Yang, Yiming and Pallier, Christophe and Hale, John},
  journal={Scientific data},
  volume={9},
  number={1},
  pages={530},
  year={2022},
  publisher={Nature Publishing Group UK London}
}

@article{nastase2021narratives,
  title={The “Narratives” fMRI dataset for evaluating models of naturalistic language comprehension},
  author={Nastase, Samuel A and Liu, Yun-Fei and Hillman, Hanna and Zadbood, Asieh and Hasenfratz, Liat and Keshavarzian, Neggin and Chen, Janice and Honey, Christopher J and Yeshurun, Yaara and Regev, Mor and others},
  journal={Scientific data},
  volume={8},
  number={1},
  pages={250},
  year={2021},
  publisher={Nature Publishing Group UK London}
}

@article{lebel2021fmri,
  title={An fMRI dataset during a passive natural language listening task},
  author={LeBel, A and Wagner, L and Jain, S and Adhikari-Desai, A and Gupta, B and Morgenthal, A and Tang, J and Xu, L and Huth, AG},
  journal={OpenNeuro. doi},
  volume={10},
  year={2021}
}

@article{toneva2019interpreting,
  title={Interpreting and improving natural-language processing (in machines) with natural language-processing (in the brain)},
  author={Toneva, Mariya and Wehbe, Leila},
  journal={Advances in neural information processing systems},
  volume={32},
  year={2019}
}

@inproceedings{
    negi2025braininformed,
    title={Brain-Informed Fine-Tuning for Improved Multilingual Understanding in Language Models},
    author={Anuja Negi and SUBBA REDDY OOTA and Anwar O Nunez-Elizalde and Manish Gupta and Fatma Deniz},
    booktitle={The Thirty-ninth Annual Conference on Neural Information Processing Systems},
    year={2025},
    url={https://openreview.net/forum?id=JPogehP8By}
}

\clearpage
\appendix

\setcounter{figure}{0}
\setcounter{table}{0}
\setcounter{page}{1}

\renewcommand{\thefigure}{S\arabic{figure}}
\renewcommand{\thetable}{S\arabic{table}}

\begin{center}
    \Large\textbf{Supplementary Information}
\end{center}

\section{Full component analysis results}

\begin{figure}[h!]
    \centering
\includegraphics[width=0.82\textwidth]{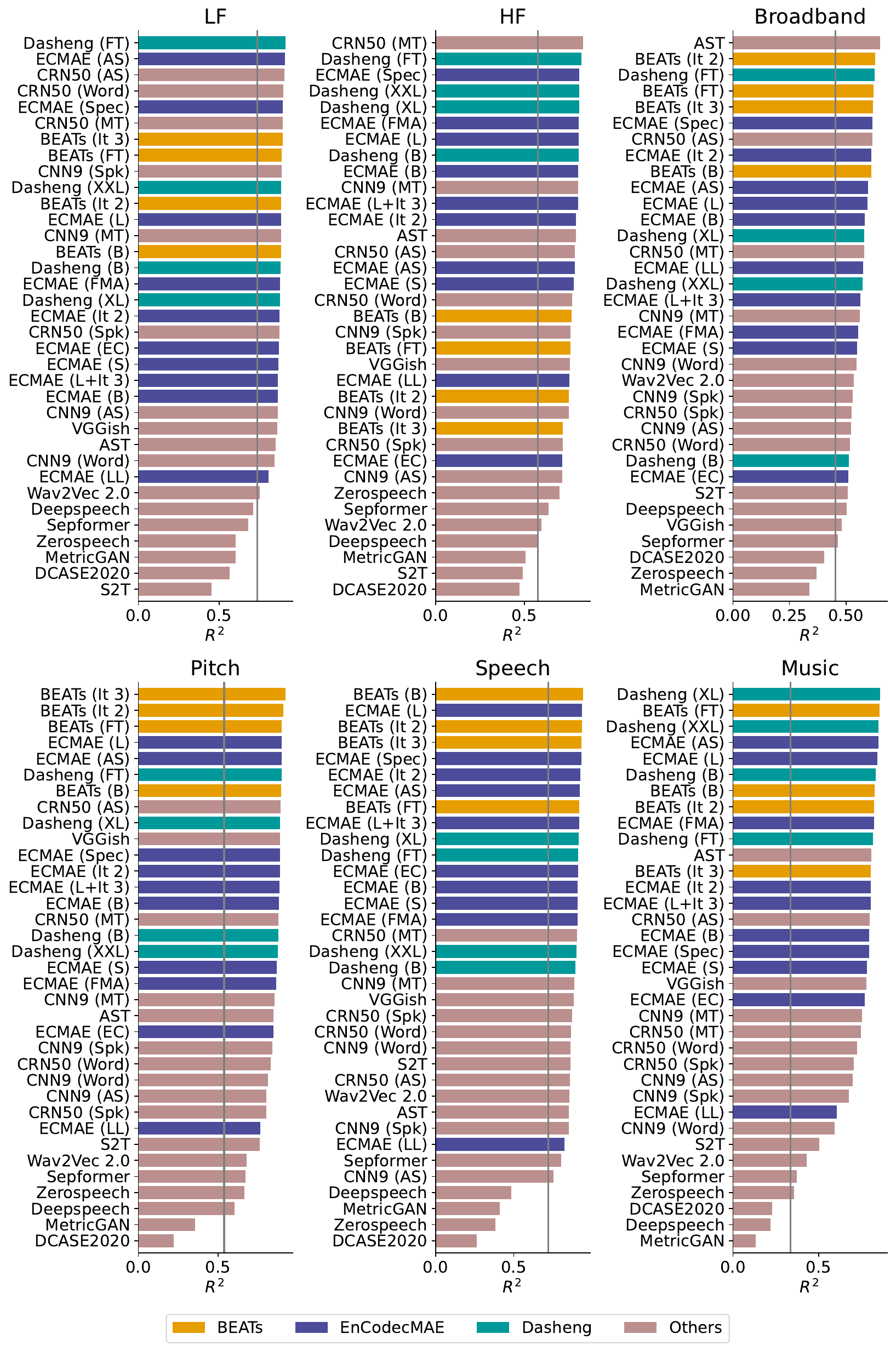}
    \caption{$R^2$ values for each of the six components and the 36 evaluated audio models. The gray lines correspond to the spectro-temporal baseline.}
    \label{fig:supp-1}
\end{figure}

\section{RSA layer selection method}

RSA calculation results in a coefficient $\rho_{m,l,s}$ for every subject $s$ and every layer $l$ of the audio model $m$, which reflects the similarity between the representations obtained from the fMRI of subject $s$ and the representations obtained from the layer $l$ of audio model $m$. We explored two approaches for obtaining a single similarity measure for the model $m$:
\begin{itemize}
    \item \textbf{Optimal layer on all data}: we computed the average of $\rho_{m,l,s}$ across subjects, leading to a Spearman correlation coefficient for each layer $\rho_{m,l}$. We finally computed the maximum value across layers as summary metric. This approach is simple but the layer selection is based on the same data which is used for reporting the test metric, which could lead to optimistic results.
    \item \textbf{Optimal layer chosen with cross-validation}: instead of using the $\rho_{m,l,s}$ values computed as above to select the best layer, for each subject we construct an audio RDM $D^a$ performing cross-validation, using the same folds over the stimuli as in the regression analysis. For each fold $f_k$, we fill the entries $D^a_{i,j}$ corresponding to stimuli $i \in f_k$ and $j \in f_k$, using entries from the RDM matrix corresponding to the best layer $l_{best}$. The best layer is selected using the entries of the RDM that correspond to stimulis not present in the fold $f_k$. After computing $D^a$ this way, the $\rho$ value is calculated as above based on this alternative RDM for the model. This method allows reporting a Spearman coefficient on data that was not used to select the best layer, while still using all stimuli for calculating the RDMs. The disadvantage is that the resulting RDMs can contain submatrices coming from different layer RDMs\footnote{Note that this is also the case for the regression analysis where the layer is selected in the inner cross-validation loop before pooling all folds together to compute the final metric.}. In practice, we observed that for most models, the selected layer was consistent across folds and subjects. 
\end{itemize}

\begin{figure}[h!]
    \centering
    \includegraphics[width=0.5\linewidth]{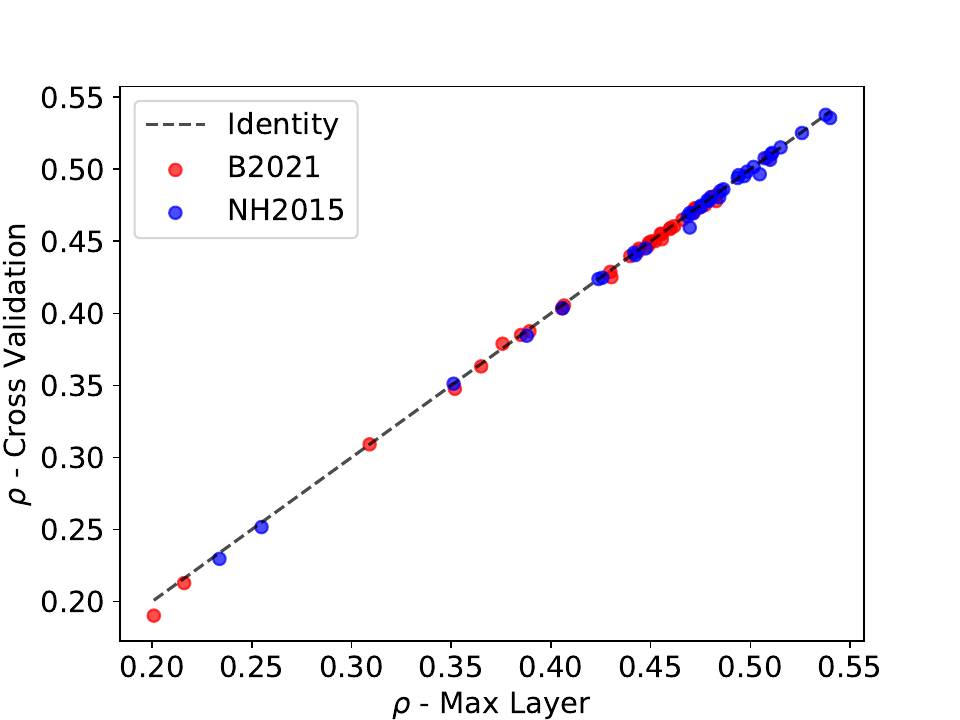}
    \caption{Comparison between the $\rho$ values obtained through RSA using cross validation and taking the maximum value across layers, for both NH2015 and B2021 datasets.}
    \label{fig:rsa-method-compare}
\end{figure}

In Figure \ref{fig:rsa-method-compare}, the results using both methods are compared. It can be seen that differences in the result are minimal, and consequently, we decided to use the simpler method. 

\end{document}